\documentclass{fairmeta}

\usepackage{amsmath}
\usepackage{amssymb}
\usepackage{enumitem}
\usepackage{booktabs}
\usepackage{multirow}
\usepackage{graphicx}
\usepackage{longtable}
\usepackage{array}
\usepackage[table]{xcolor}  % 可选：隔行底色
\usepackage{fontawesome5}
\definecolor{gold}{HTML}{D4AF37}
\definecolor{silver}{HTML}{9E9E9E}
\definecolor{bronze}{HTML}{CD7F32}
\newcommand{\medal}[2]{\textcolor{#1}{\faMedal}\,\textbf{#2}}
\usepackage{pifont,xcolor}
\newcommand{\cmark}{\textcolor{green!55!black}{\ding{51}}}
\newcommand{\xmark}{\textcolor{red!70!black}{\ding{55}}}
\usepackage{tabularx}
\setcitestyle{round,authoryear}

\DeclareMathSymbol{@}{\mathord}{letters}{"3B}
\def\name{GameHorizon}
\def\namesuite{GameHorizon Suite}
\def\numgame{$21$}
\def\nummodel{$47$}

\def\lengthgamesplit{$5@000$ hours}
\def\data{GameHorizon-Data}
\def\anno{GameHorizon-Annotator}
\def\bench{GameHorizon-Bench}

\def\reffig{Fig.}
\def\reftab{Table}

\def\refsec{Sec.}

\title{\Large GameHorizon Suite: Multi-Horizon Data and Evaluation in Gameplay}

\author[1,*,\dagger]{Yiran Wang}
\author[1,2,6,*]{Xingyilang Yin}
\author[1,*]{Junfu Pu}
\author[1,*]{Guangzhi Wang}
\author[2]{Kaifeng Li}

\author[1,3]{Mingyu Ouyang}
\author[1,4]{Huiqiang Sun}
\author[1,5]{Lingen Li}
\author[1]{Cheng Cheng}
\author[1]{Wangbo Yu}
\author[1]{Honghao Chen}
\author[1,2,\text{\ding{41}}]{Xiaodong Cun}
\author[6]{Chi-Man Pun}
\author[4]{Zhiguo Cao}
\author[1]{Ying Shan}
\affiliation[1]{ARC Lab, Tencent IEG}
\affiliation[2]{GVC Lab, Great Bay University}
\affiliation[3]{National University of Singapore}
\affiliation[4]{\mbox{Huazhong University of Science and Technology}}
\affiliation[5]{MMLab, CUHK}
\affiliation[6]{University of Macau}
\contribution[\dagger]{Project Lead}
\contribution[*]{Equal Contribution}
\contribution[\text{\ding{41}}]{Corresponding Author}

\abstract{Modern video games provide a measurable testbed for AI models, combining abilities of visual understanding, instruction decomposition, goal planning, and precise action control over multiple temporal horizons. Existing datasets and benchmarks, however, either cover a narrow range of games, lack language instructions, or rely on high-variance online rollouts. To address these challenges, we introduce \name{}, a unified data and evaluation suite that measures gameplay capabilities at different horizons for diverse model families. \namesuite{} consists of three components. First, \anno{} is a scalable and automated annotation pipeline for multi-horizon instructions. Second, utilizing the pipeline, we construct \data{}, the first large-scale AAA gameplay dataset with temporally aligned videos, player actions, and multi-horizon instructions. It comprises \lengthgamesplit{} of recordings from \numgame{} games, collected by $100$ human expert players. Third, we build \bench{} with reproducible offline and stepwise online testing. The offline track enables reproducible evaluation using thousands of standardized questions organized into three primary tasks and a series of diagnostic variants, while the online track tests whether offline scores reflect actual gameplay capabilities and localizes failures to specific steps within long-horizon gameplay. Based on our \namesuite{}, we evaluate \nummodel{} models through more than one million model invocations, revealing a meaningful hierarchy of task difficulty and pronounced differences in model capabilities. Our work can provide a standardized yardstick for evaluating gameplay capabilities across horizons and model families. We will release our dataset, annotator, and benchmark to facilitate future research. \vspace{-20pt}
}

\metadata[Project Page]{\url{https://gamehorizon-suite.github.io}}
\metadata[Github]{\url{https://github.com/TencentARC/GameHorizon}}
\begin{document}

\maketitle

\begin{figure}
    \centering
    %\vspace{-10pt}  
    \includegraphics[width=1.0\textwidth,trim=20 0 5 5,clip]{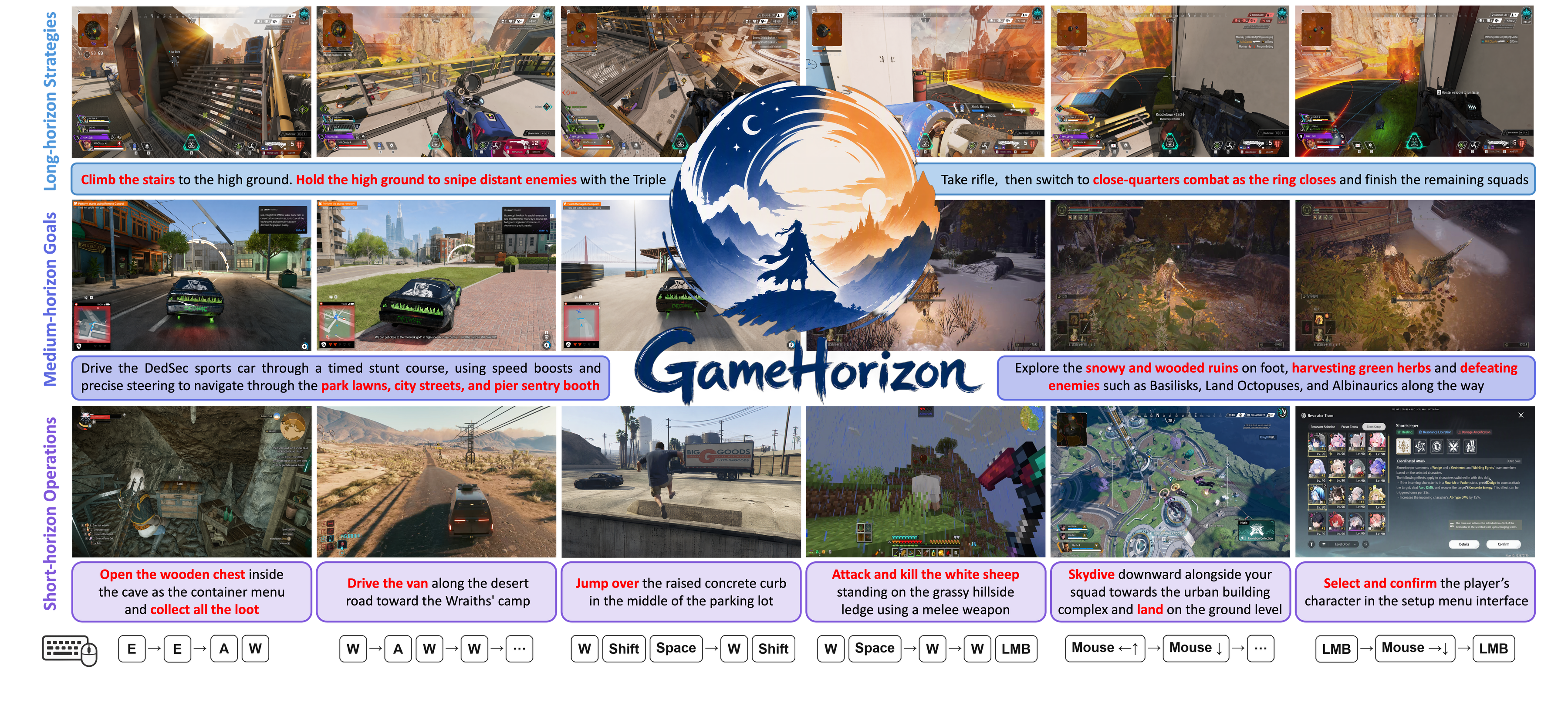}
    \vspace{-30pt}
    \caption{\textbf{\namesuite{}} spans diverse games with a pyramid of primitive actions (\textit{e.g.}, Left Mouse Button as LMB), short-horizon operations, medium-horizon goals, and long-horizon strategies, providing a unified yardstick for testing gameplay capabilities across models and temporal horizons. Red text highlights key actions or objects in video frames.}

  \label{fig:fig1_teaser} 
  \vspace{-7pt}
\end{figure}

\section{Introduction}
\label{sec:intro}

Empowering AI models to play modern video games provides a measurable testbed for understanding, decision-making, and acting within complex environments. Game objectives span varying temporal scales, \textit{e.g.}, collecting an item within seconds, winning a fight lasting several minutes, and executing a strategy that unfolds over an entire game session, as shown in \reffig{}~\ref{fig:fig1_teaser}. Whatever the horizon, these objectives are all reflected in the same stream of primitive actions, such as keystrokes and mouse movements. Thus, a growing number of dedicated game agents~\citep{nitrogen,openp2p,lumine,gametars,minestudio} have emerged. General-purpose vision-language models (VLMs) and agents~\citep{gemini25,gemini1,gpt4,vpt,sima2, uitars1,uitars2,voyager,cradle} have also begun to treat video games as a capability target, \textit{e.g.}, SIMA~2~\citep{sima2} equips Gemini~\citep{gemini25} to follow instructions in open-world games.

Playing a game well requires two abilities at once: (1) based on an understanding of the situation so far, planning subsequent behavior by decomposing the long-horizon game objective into subgoals; and (2) turning the plans and goals into concrete actions. In practice, these two abilities are split across two families of models. Most dedicated game agents~\citep{nitrogen,openp2p,jarvisvla,groot,rocket1} are optimized for acting rather than planning. To sustain a high control frequency, they adopt lightweight vision-language-action (VLA) or action-head architectures~\citep{RT2,openvla} under action-trajectory supervision, at the cost of the capacity to reason and plan. Conversely, general-purpose models~\citep{gemini25,gemini1,gpt4,claude37} excel at planning, yet they have never been systematically measured on action. The few reported cases, \textit{e.g.}, Gemini and Claude playing Pok\'emon~\citep{gemini25,claude37}, rely on bespoke agent harnesses and cannot be compared across models. What is needed, therefore, is a single yardstick for both families: one that measures whether a model can align vision, executable actions, and multi-horizon natural-language goals. To be useful, the yardstick should be low-cost, standardized, and reproducible, independent of harnesses or environments. 

Existing game datasets and benchmarks fall short of these requirements. First, constrained by annotation cost, their game coverage is narrow. For instance, GameWorld~\citep{gameworld} targets simple mini-games. STEVE-1~\citep{steve1} and MineDojo~\citep{minedojo} are confined to Minecraft. WildWorld~\citep{wildworld} is collected from a single game, Monster Hunter Wilds. Conclusions drawn from a single title or simplified mini-games cannot generalize to complex and heterogeneous AAA games. Second, human–agent interaction is largely mediated through language, \textit{e.g.}, instruction following, planning, and goal decomposition. However, previous attempts still lack comprehensive annotations of text instructions and goals. NitroGen~\citep{nitrogen} and GameVerse~\citep{gameverse} omit instructions entirely, while Open-P2P~\citep{openp2p} only provides highly sparse annotations. Consequently, existing data and benchmarks cannot systematically evaluate model performance across instruction following, goal planning, and action execution. Third, prior work mainly relies on online evaluations with a limited number of agent rollouts. Lumine~\citep{lumine} reports task success rates only by three trials per scene, while GameVerse~\citep{gameverse} conducts rollouts on $3$--$20$ cases. Such small samples lead to low-confidence comparisons. Online results are sensitive to specific game environments and agent harnesses, making them difficult to reproduce. Moreover, an aggregate success rate collapses distinct failure modes into a single scalar. When a model fails, it remains unclear whether it misidentifies current actions, infers the next goal incorrectly, or fails to map the goal to right future controls.

{\parfillskip=0pt
To address these challenges, as presented in \reffig{}~\ref{fig:fig1_teaser}, we introduce \textbf{\name{}}, a data and evaluation suite spanning multiple temporal horizons and AAA games. It serves as a unified yardstick across a broad range of model types. Specifically, \namesuite{} consists of three key components. First, \textbf{\anno{}} is a scalable annotation pipeline for multi-horizon instructions in gameplay. Unlike the manual annotation in Game-TARS~\citep{gametars}, \anno{} automatically produces a pyramid of natural-language instructions at three temporal horizons, including short-horizon operations, medium-horizon goals, and long-horizon strategies. The pipeline operates bottom-up, abstracting fine-grained instructions into higher levels. Second, based on the annotator, we construct \textbf{\data{}}, a large-scale gameplay dataset with \lengthgamesplit{} of recordings collected from $100$ human expert players across \numgame{} game titles. \data{} is the first publicly available dataset that aligns game frames, player actions, and multi-horizon instructions, which also exceeds previous corpora such as D2E~\citep{d2e} and gaming-500-hours~\citep{gaming500h} in scale. Third, we build \textbf{\bench{}}, combining reproducible offline evaluation and stepwise online testing. The offline track comprises three sorts of primary tasks: single-horizon action, multi-horizon instruction decomposition, and cross-horizon consistency. Additional variants enable model diagnosis at a finer granularity. The offline track is reliable and reproducible based on thousands of questions with standardized actions and instructions. Besides, online track evaluates long-horizon gameplay through short-horizon subtasks, covering order-dependent causal tasks and order-flexible thematic tasks. Environment reset enables stepwise verification and failure localization. This track tests whether offline scores reflect actual gameplay abilities.\par
}

%The online track measures gameplay for step-wise verification, localizing failures to specific stages by environment resetting. We design long-horizon tasks composed of short-horizon subtasks, \textit{e.g.}, causal tasks requiring sequential step completion and thematic tasks allowing varying order. The online track examines whether offline scores reflect actual gameplay ability.

%We design $20$ long-horizon tasks, either causal, where each step depends on the previous ones, or thematic, where the steps can be completed in any order, together with the $50$ short-horizon subtasks they comprise. We localize to specific stages

%A checkpoint mechanism resets the environment to the state in which a failed subtask has been completed, so that the remaining steps are still evaluated and failures are localized to specific stages rather than collapsed into a single success rate. The online track examines whether offline scores reflect actual gameplay ability and tests whether instructions at longer horizons help models act in a live environment.

{\parfillskip=0pt
We conduct extensive empirical evaluations based on the \data{} and \bench{}. Our data covers various game categories, \textit{e.g.}, open-world, action role-playing, competitive shooter, sandbox survival, and creature-collecting adventure genres. Our benchmark involves more than one million model inferences and API calls. For the offline setting, we test \nummodel{} models on our primary tasks, including general-purpose VLMs~\citep{gemini25,gpt4,kimik3,qwen3}, unified multimodal models (UMMs)~\citep{sensenova,ovis,internvl-u,bagel}, coding and GUI agents~\citep{uitars1,uitars2,stepgui,claude}, as well as dedicated game agents~\citep{gametars,openp2p,nitrogen,jarvisvla}. The results reveal a meaningful hierarchy of task difficulty and pronounced differences in model capabilities. For the online setting, we observe a clear positive association between task success rates and offline scores, suggesting that the offline accuracy provides a valid proxy for actual gameplay capabilities. Beyond aggregate performance, we further analyze the bottlenecks of current models. Planning future actions and decomposing complex goals are more challenging than deciding the current action. Compared with the vision-only input, incorporating our medium- and long-horizon instructions improves future-action planning by $7.2$ percentage points, highlighting the effectiveness of our multi-horizon instructions. Our main contributions can be summarized as follows:\par
}

\begin{itemize}[leftmargin=*]

    \item We introduce \name{}, a data and evaluation suite with multiple  temporal horizons and AAA games.

    \item \anno{} works as a scalable annotation pipeline for multi-horizon instructions in gameplay.
    
    \item \data{} is a large-scale gameplay dataset with aligned triplets of videos, actions, and instructions.

    \item \bench{} unifies reproducible offline and stepwise online evaluations for diverse model families.

\end{itemize}

\section{Related Work}

\noindent \textbf{Gameplay Datasets and Benchmarks.} Prior gameplay data and benchmarks suffer from narrow game coverage, limited instruction annotations, and high-variance evaluations. First, many datasets are confined to a single game or simplified mini-games. WildWorld~\citep{wildworld} is collected from Monster Hunter Wilds. MineDojo~\citep{minedojo}, VPT~\citep{vpt}, STEVE-1~\citep{steve1}, MineRL~\citep{minerl}, and MCU~\citep{mcu} provide data and evaluation exclusively in Minecraft. Some efforts attempt to encompass multiple titles. However, constrained by annotation costs, they either remain limited in scale (\textit{e.g.}, 300 hours for D2E by \citealp{d2e} and 500 hours for gaming-500-hours by \citealp{gaming500h}) or fall back on mini-games (\textit{e.g.}, GameWorld by \citealp{gameworld}). Second, existing corpora lack comprehensive text instruction annotations, which are critical for human–agent interaction, \textit{e.g.}, instruction following and goal planning. NitroGen~\citep{nitrogen}, GameVerse~\citep{gameverse}, D2E~\citep{d2e}, and VPT~\citep{vpt} are entirely devoid of language instructions, whereas Open-P2P~\citep{openp2p} only offers sparse and coarse annotations. Game-TARS~\citep{gametars} relies on costly manual annotations, which hinders scalability and remains unreleased. Third, previous benchmarks mainly adopt online evaluations with a limited number of rollouts. Lumine~\citep{lumine} reports success rates by three trials per scene. GameVerse~\citep{gameverse} conducts rollouts on $3$--$20$ cases. VideoGameBench~\citep{videogamebench} likewise tests each model with a single run per game. Such small sample sizes lead to low-confidence comparisons. Online results are sensitive to game environments and custom harnesses, making them difficult to reproduce. In contrast, \name{} provides a unified data and evaluation suite, featuring \lengthgamesplit{} of recordings, diverse AAA game genres, multi-horizon instructions, and reproducible offline-online benchmarks.

%However, constrained by annotation costs, they either remain limited in scale (\textit{e.g.}, $300$ hours for D2E~\citep{d2e} and $500$ hours for gaming-500-hours~\citep{gaming500h}) or fall back on simplified mini-games (\textit{e.g.}, GameWorld~\citep{gameworld}). Second, most existing corpora lack comprehensive annotations of text instructions, which are critical for human–agent interaction, \textit{e.g.}, instruction following and goal planning. For instance, NitroGen~\citep{nitrogen}, GameVerse~\citep{gameverse}, D2E~\citep{d2e}, and VPT~\citep{vpt} are entirely devoid of language instructions, whereas Open-P2P~\citep{openp2p} only offers sparse and coarse annotations. Game-TARS~\citep{gametars} relies on costly manual annotations, which hinders scalability and remains unreleased.

% 
% Second, most existing corpora lack comprehensive language annotations. Large video--action collections often omit instructions entirely~\citep{nitrogen,gameverse,d2e,gaming500h}, while Open-P2P~\citep{openp2p} provides only sparse text labels. Even when language is available, it is typically obtained through costly manual annotation and does not form a consistent hierarchy across short-horizon operations, medium-horizon goals, and long-horizon strategies~\citep{gametars}. Without temporally aligned multi-horizon instructions, instruction following, goal decomposition, and long-horizon planning cannot be evaluated systematically.

\noindent \textbf{Game-playing Models.}
Video games serve as a practical testbed for AI models to perceive, plan, decide, and act in complex environments. Various studies leverage games to enhance or test model capabilities. On one hand, dedicated game agents are typically tailored for high-frequency action control, often at the expense of reasoning abilities, \textit{e.g.}, long-horizon goal decomposition and planning. Open-P2P~\citep{openp2p} employs an EfficientNet~\citep{efficientnet} as a visual encoder alongside a lightweight action decoder for low-latency inference on consumer GPUs, whereas JARVIS-VLA~\citep{jarvisvla} instantiates a VLA policy with a short context window. On the other hand, general-purpose models, \textit{e.g.}, VLMs, UMMs, and computer-use agents, exhibit stronger cognitive and planning capabilities, yet they have not been systematically evaluated on action execution. Cradle~\citep{cradle} couples GPT-4V~\citep{gpt4} with a multi-module agent harness on commercial games. Gemini~\citep{gemini25} and Claude~\citep{claude} have been tested on \textit{Pokémon} through bespoke agent loops. These evaluations remain incomparable due to specialized setups and harnesses. To bridge this gap, GameHorizon presents a unified and standardized yardstick for different models, enabling evaluations across goal planning, instruction following, and executable actions at multiple horizons.

%Dedicated game agents such as VPT~\citep{vpt}, STEVE-1~\citep{steve1}, JARVIS-VLA~\citep{jarvisvla}, NitroGen~\citep{nitrogen}, Open-P2P~\citep{openp2p}, Lumine~\citep{lumine}, and Game-TARS~\citep{gametars} are typically trained with action-trajectory supervision and lightweight vision--language--action or action-head architectures.
%This design supports high-frequency control, but often at the expense of explicit reasoning, instruction decomposition, and long-horizon planning.
%A complementary line of work applies general-purpose models to games.
%Vision-language models and agents have been tested in open-world and commercial settings, \textit{e.g.}, SIMA 2~\citep{sima2} with Gemini, Voyager~\citep{voyager} in Minecraft, and Cradle~\citep{cradle} on AAA titles; coding and GUI agents such as UI-TARS~\citep{uitars1,uitars2} likewise treat gameplay as a computer-use problem.
%These models exhibit stronger planning and language-conditioned reasoning, yet they are rarely evaluated on the same executable action interface or across multiple temporal horizons.
%In contrast, \name{} provides a unified yardstick for both families, measuring whether models can align visual observations, keyboard-and-mouse actions, and multi-horizon instructions under reproducible offline and online protocols.
\section{\namesuite{}}

% v6单卡 v8金字塔Logo  v7叠叠乐不好 v9anno加粗  v10图例加粗
\begin{figure}[!t]
    \centering
    \vspace{-1pt}  
    \includegraphics[width=1.0\textwidth,trim=20.5 0 4 12,clip]{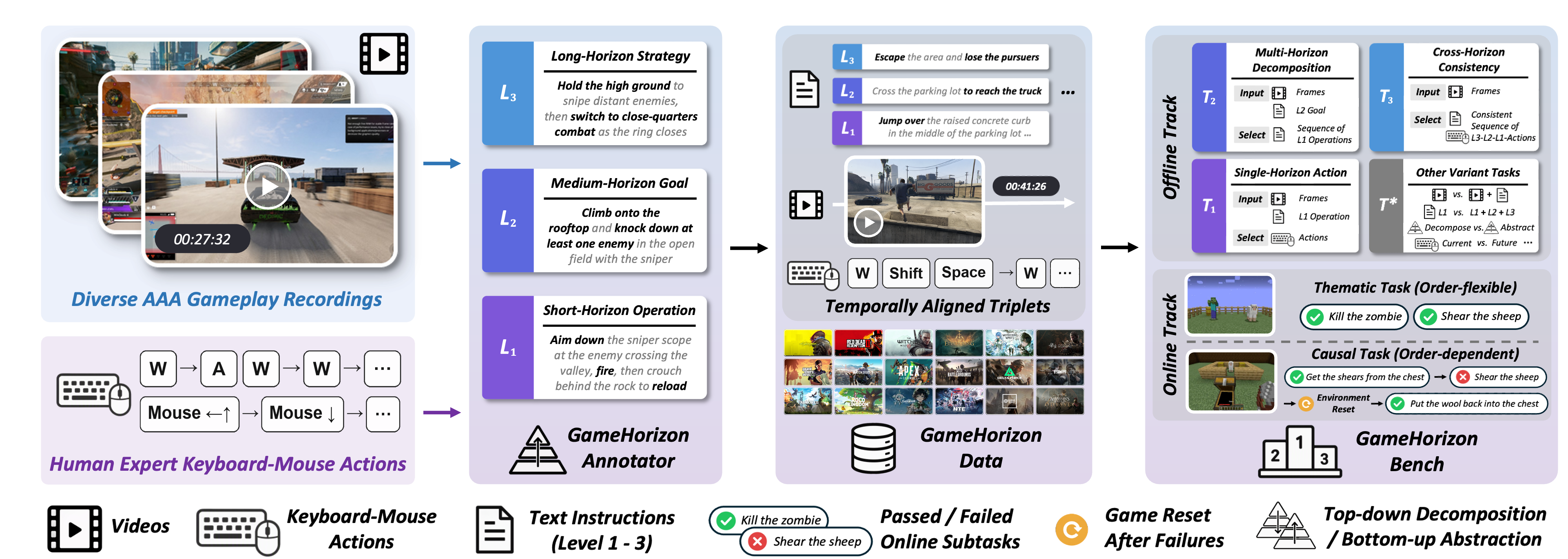}
    \vspace{-14pt}
    \caption{\textbf{Overview of \namesuite{}.} 
    %Our framework comprises three components: the Annotator, Data, and Bench.
    %Our framework comprises \anno{}, -Data, and -Bench. %We collect $5@000$ hours of synchronized gameplay videos and keyboard-mouse actions. %Given the recordings and action trajectories, 
    \anno{} automatically produces %gameplay videos and actions with 
    a three-level pyramid of %textual instructions, \textit{i.e.}, 
    short-horizon operations $L_1$, medium-horizon goals $L_2$, and long-horizon strategies $L_3$. \data{} contains $5@000$ hours of gameplay across $21$ game titles, with temporally aligned videos, actions, and multi-horizon instructions. \bench{} provides reproducible offline and stepwise online testing. The offline track contains thousands of standardized questions across three primary tasks $T_1$–$T_3$ and diagnostic variants $T^*$. The online track tests order-dependent causal and order-flexible thematic tasks through verifiable subtasks and game resets for failure localization.}

  \label{fig:pipeline} 
  \vspace{-7pt}
\end{figure}

We present \name{}, a unified data and evaluation suite spanning multiple temporal horizons, diverse AAA games, and a broad range of model families. We outline our approach in \refsec{}~\ref{sec:3.1}. In \refsec{}~\ref{sec:3.2}, we elaborate on \anno{}, an automated and scalable annotation pipeline for multi-horizon instructions. Our large-scale \data{} is discussed in Sec.~\ref{sec:3.3}, while \bench{} is illustrated in \refsec{}~\ref{sec:3.4}.

\subsection{Overview} \label{sec:3.1} 

As shown in \reffig{}~\ref{fig:pipeline}, \namesuite{} comprises three components: \anno{}, \data{}, and \bench{}. The workflow begins with raw gameplay acquisition. Previous work collects web videos and recovers action pseudo-labels via Inverse Dynamics Models (IDMs)~\citep{vpt,steve1,d2e} or gamepad segmentation~\citep{nitrogen,segformer}. The inferred pseudo-labels can deviate from the actual controls executed by humans. In contrast, we recruit $100$ experienced human players and deploy a dedicated recording-and-upload system to synchronously capture game videos at 2K resolution along with timestamped keyboard and mouse actions. Authentic human gameplay recordings and action trajectories can establish a reliable foundation for faithful evaluations in complex game worlds.

Based on the recordings, we develop \anno{}, an annotation pipeline for textual instructions across multiple temporal horizons. Due to annotation costs or vision-only architectures, 
prior studies either omit instructions~\citep{nitrogen,gameverse} or provide sparse labels~\citep{openp2p}. Game-TARS~\citep{gametars} relies on manual instruction labeling, which is expensive and remains unavailable to the community. However, instructions are vital for human-agent interaction, \textit{e.g.}, instruction following and goal decomposition. When issuing requests to an agent, humans dictate not only primitive actions like reloading a weapon, but also long-term objectives such as defending a bridge. To this end, our \anno{} produces a three-level pyramid of instructions, including short-horizon operations ($1$--$5$ seconds), medium-horizon goals ($1$--$2$ minutes), and long-horizon strategies ($5$--$8$ minutes). The annotator operates bottom-up, abstracting dense and action-grounded instructions into higher-level goals and strategies. The automated pipeline reduces annotation costs and enables scalable labeling across large gameplay collections.

Applying \anno{} to the collected trajectories, we construct \data{}, the first large-scale AAA gameplay dataset that aligns videos, player actions, and multi-horizon instructions. It covers $5@000$ hours of human gameplay from $21$ game titles, spanning diverse genres such as open-world, action role-playing, competitive shooter, sandbox survival, and creature-collecting adventure titles. In total, \data{} contains $4@571$ videos recorded at $60$ fps and $411.03$ million keyboard-mouse action events. It is annotated with $6@184@036$ distinct instructions, including $5@947@588$ short-horizon operations, $189@158$ medium-horizon goals, and $47@290$ long-horizon strategies. On average, \data{} provides one distinct short-horizon instruction every $2.63$ seconds, while each frame is aligned with corresponding instructions at all three horizons. Our annotations are substantially denser than those in prior work. For example, Open-P2P~\citep{openp2p} includes only one instruction every few minutes, with uneven temporal coverage. Combining scale, density, and diversity, \data{} enables unified evaluations of multi-horizon gameplay tasks and capabilities.

Finally, leveraging \data{}, we introduce \bench{}, a comprehensive benchmark featuring reproducible offline and stepwise online testing across diverse model families. As discussed in \refsec{}~\ref{sec:intro}, prior gameplay benchmarks predominantly rely on a small number of harness-dependent online rollouts~\citep{videogamebench,lumine,gameverse}, yielding low-confidence comparisons that are difficult to reproduce. Aggregate success rates also conflate different failure modes. In contrast, the offline track of our \bench{} ensures reliable and reproducible evaluation through thousands of multiple-choice questions (MCQs) with standardized actions and instructions in three primary tasks, including single-horizon action, multi-horizon instruction decomposition, and cross-horizon consistency. Additional diagnostic variants further probe model capabilities along different dimensions, \textit{e.g.}, current-action perception \textit{vs.} future-action planning and top-down decomposition \textit{vs.} bottom-up abstraction. Complementarily, the online track evaluates long-horizon gameplay through collections of verifiable short-horizon subtasks. It covers causal tasks, whose subtasks follow a prescribed sequence, and thematic tasks, whose subtasks can be completed in any order. When an agent fails at a subtask, the environment is reset to the corresponding success state, allowing evaluation to continue. The stepwise protocol localizes errors to specific steps. Our online track further tests whether offline scores reflect actual gameplay abilities. Together, the two tracks establish \bench{} as a unified, standardized, reproducible, and diagnostic yardstick across model families and temporal horizons.

%Third, we build \textbf{\bench{}}, combining reproducible offline evaluation and stepwise online testing. The offline track comprises three sorts of primary tasks: single-horizon action, multi-horizon instruction decomposition, and cross-horizon consistency. Additional variants enable model diagnosis at a finer granularity. The offline track is reliable and reproducible based on thousands of questions with standardized actions and instructions. Besides, online track evaluates long-horizon gameplay through short-horizon subtasks, covering order-dependent causal tasks and order-flexible thematic tasks. Environment reset enables stepwise verification and failure localization. This track tests whether offline scores reflect actual gameplay abilities.

\begin{figure}[!t]
    \centering
    \vspace{-1pt}  
    \hspace{+4pt}\includegraphics[width=0.98\textwidth,trim=0 0 6 0,clip]{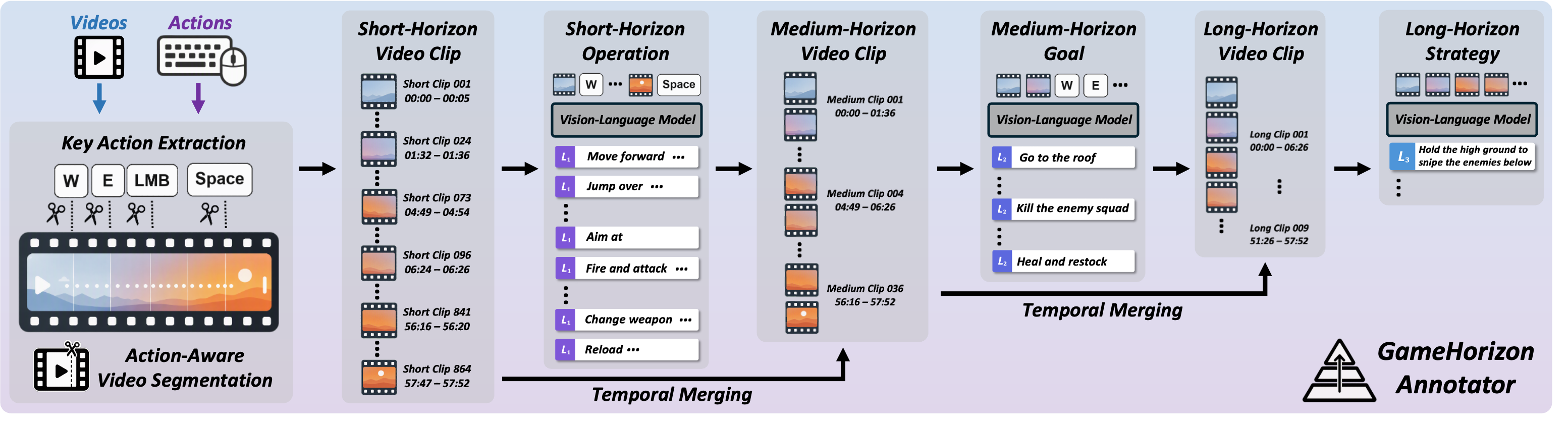}
    \vspace{-5pt}
    \caption{\textbf{Workflow of \anno{}.} 
    Videos and actions are first processed by action-aware segmentation to produce short-horizon clips, with key actions determining their temporal boundaries. A VLM annotates each clip with an $L_1$ operation. Lower-level clips with their instructions are progressively merged into medium- and long-horizon clips based on action continuity and semantic coherence, from which the VLM derives $L_2$ goals and $L_3$ strategies. The bottom-up procedure abstracts fine-grained video-action trajectories into a pyramid of multi-horizon text instructions.} 

  \label{fig:anno-pipe} 
  \vspace{-7pt}
\end{figure}

\subsection{\anno{}} \label{sec:3.2}

%Given a synchronized gameplay video $\mathbf V$ and action stream $\mathbf A$, \anno{} constructs a three-level pyramid of temporally aligned video clips, action traces, and textual instructions. At level $l\in{1,2,3}$, the $i$-th segment is represented as

%Given a synchronized gameplay video $\mathbf V$ and action stream $\mathbf A$, \anno{} constructs a three-level pyramid $\mathcal H=(\mathcal H_1,\mathcal H_2,\mathcal H_3)$. Each horizon is represented as a collection of temporally aligned triplets:

%\begin{equation}
%\mathcal{H}_h =
%\{\mathbf{V}^i_h,\mathbf{A}^i_h,L^i_h\}_{i=1}^{N_h},\;h\in\{1,2,3\}\,,
%\label{eq:1}
%\end{equation}

%where $N_h$ denotes the number of segments at horizon $h$, and $\mathbf V_i^h$, $\mathbf A_i^h$, and $L_i^h$ denote the video clip, aligned action trace, and textual instruction of the $i$-th segment, respectively. The three horizons correspond to short-horizon operations, medium-horizon goals, and long-horizon strategies.

Given synchronized videos and actions, \anno{} constructs a three-level instruction pyramid, \textit{i.e.}, short-horizon operations, medium-horizon goals, and long-horizon strategies. As illustrated in \reffig{}~\ref{fig:anno-pipe}, our workflow comprises the action-aware segmentation, bottom-up temporal merging, and instruction annotation. 

\noindent \textbf{Action-Aware Video Segmentation.} Constructing multi-horizon instructions from videos in a top-down manner is challenging, as VLMs struggle to resolve fine-grained visual and action details across extended temporal contexts, \textit{e.g.}, an hour-long gameplay session. We therefore proceed bottom-up, partitioning long videos into short clips that VLMs can interpret more reliably. However, off-the-shelf segmentation tools such as PySceneDetect~\citep{pyscene} rely on frame-to-frame visual similarity and tend to over-segment continuous actions, \textit{e.g.}, under rapid camera motion or abrupt viewpoint shifts. To address this issue, we perform action-aware segmentation using keyboard-mouse traces to identify key action transitions and determine clip boundaries. Specifically, we map raw keyboard-mouse events to game-specific action semantics, \textit{e.g.}, \texttt{Shift} as sprinting in Cyberpunk 2077. We then scan the mapped action stream chronologically, grouping consecutive actions within the same sustained event, such as alternating between walking and running during a single traversal. Discrete action events such as jumping or attacking define initial boundaries for coarse segmentation. Since the same keyboard-mouse input may carry different action semantics, \textit{e.g.}, a left click may indicate an attack or item selection, we employ a VLM~\citep{gemini35flash} to examine video frames, disambiguate action semantics, and refine the initial segmentation to a final set of short-horizon clips.

\noindent \textbf{Bottom-Up Temporal Merging.} Based on the short-horizon ($L_1$) clips and instructions, we merge adjacent segments into medium-horizon ($L_2$) clips. Specifically, the VLM~\citep{gemini35flash} determines whether neighboring $L_1$ operations form a continuous progression toward the same medium-horizon goal. Similarly, to construct long-horizon ($L_3$) clips, the VLM assesses whether neighboring $L_2$ goals follow the same gameplay strategy. Besides, we apply a dynamic programming algorithm~\citep{bellman1966dynamic} to enforce level-specific duration ranges of $1$--$5$ seconds for $L_1$, $1$--$2$ minutes for $L_2$, and $5$--$8$ minutes for $L_3$. The short range helps avoid fragmenting primitive actions, while the long range keeps clip durations within the reasoning capacity of current models. We term the temporal merging process bottom-up because it starts from numerous fine-grained $L_1$ clips at the base of the pyramid and progressively merges them into fewer higher-level clips. 

\noindent \textbf{Multi-Horizon Instruction Annotation.} The VLM~\citep{gemini35flash} produces instructions at each level using specific prompts with distinct input modalities and output granularities. For an $L_1$ clip, the VLM receives sampled video frames and temporally aligned keyboard-mouse actions as input. The prompt asks the VLM to generate an action-grounded instruction describing the current operation of the player. We also require the VLM to include sufficient details, such as coordinates, object descriptions, and spatial relations, when needed to guide a virtual agent in game sessions. For an $L_2$ clip, frames, actions, and the constituent $L_1$ instructions serve as inputs to the VLM. The VLM synthesizes these inputs into a medium-horizon instruction describing the goal pursued by the player. We require the output to retain only necessary details while avoiding excessive local information. For an $L_3$ clip, the inputs comprise video frames and the constituent $L_2$ instructions, without action traces. The VLM generates a long-horizon instruction that captures the high-level gameplay strategy. The prompt emphasizes the long-term intent while suppressing local operational details.

For $L_1$ and $L_2$, the action inputs to the VLM ground the instructions in controls executed by players rather than visual evidence alone. The actions are useful for identifying visually ambiguous operations, \textit{e.g.}, determining whether the player detaches from the squad during skydiving in Apex Legends. By comparison, we omit actions at $L_3$ to keep the instructions focused on high-level strategies rather than detailed operations. 

All prompts include the game title, clip duration, frame rate, and sampled frame indices for reference. At higher levels, the constituent lower-level instructions provide compact semantic context and help maintain consistency across horizons. The $L_1$ and $L_2$ prompts also incorporate game-specific keybinds that map keyboard-mouse actions to corresponding in-game semantics, facilitating action interpretation by the VLM.

\begin{table*}[!t]
\centering
\caption{\textbf{Statistics of \data{} across $\boldsymbol{21}$ game titles.} We report the duration and share of recordings, counts of actions and instructions, and the average temporal span of each distinct instruction. Games are sorted by duration. Valid rate denotes the fraction retained for instruction annotation after filtering. Actions are reported in units of $10^4$.}
\vspace{-7pt}
\label{tab:data_statistics}
\setlength{\tabcolsep}{3.5pt}
\renewcommand{\arraystretch}{1.08}
\resizebox{\textwidth}{!}{%
\begin{tabular}{l*{11}{c}}
\toprule
\multirow{2.5}{*}{\textbf{Game Title}} &
\multicolumn{4}{c}{\textbf{Recording Duration and Share}} &
\multicolumn{4}{c}{\textbf{Annotation Counts}} &
\multicolumn{3}{c}{\textbf{Avg. Instruction Span (s)}} \\
\cmidrule(lr){2-5}
\cmidrule(lr){6-9}
\cmidrule(lr){10-12}
&
\textbf{Total (h)} &
\textbf{Share (\%)} &
\textbf{Valid Rate (\%)} &
\textbf{Valid (h)} &
\textbf{Actions ($\boldsymbol{10^4}$)} &
$\boldsymbol{L_1}$ &
$\boldsymbol{L_2}$ &
$\boldsymbol{L_3}$ &
\makebox[1cm][c]{$\boldsymbol{L_1}$} &
\makebox[1cm][c]{$\boldsymbol{L_2}$} &
\makebox[1cm][c]{$\boldsymbol{L_3}$} \\
\midrule

Valorant
& $617.5$ & $12.35$ & $86.4$ & $533.8$ & $5@811$
& $705@380$ & $23@113$ & $5@778$ & $2.72$ & $83.1$ & $332.5$ \\

Minecraft
& $589.8$ & $11.80$ & $84.5$ & $498.3$ & $5@087$
& $448@166$ & $21@715$ & $5@429$ & $4.00$ & $82.6$ & $330.4$ \\

Grand Theft Auto V
& $566.1$ & $11.32$ & $86.5$ & $489.9$ & $3@585$
& $382@302$ & $21@644$ & $5@411$ & $4.61$ & $81.5$ & $325.9$ \\

Palworld
& $438.7$ & $8.77$ & $83.5$ & $366.5$ & $3@741$
& $329@615$ & $15@971$ & $3@993$ & $4.00$ & $82.6$ & $330.4$ \\

Delta Force
& $337.0$ & $6.74$ & $85.1$ & $286.9$ & $3@123$
& $379@091$ & $12@422$ & $3@105$ & $2.72$ & $83.1$ & $332.5$ \\

Roco Kingdom: World
& $329.2$ & $6.58$ & $91.9$ & $302.4$ & $2@677$
& $786@446$ & $13@054$ & $3@263$ & $1.38$ & $83.4$ & $333.6$ \\

Red Dead Redemption 2
& $304.4$ & $6.09$ & $85.9$ & $261.6$ & $1@992$
& $202@428$ & $11@553$ & $2@888$ & $4.65$ & $81.5$ & $326.1$ \\

Cyberpunk 2077
& $301.2$ & $6.02$ & $87.9$ & $264.7$ & $2@574$
& $192@222$ & $11@646$ & $2@912$ & $4.96$ & $81.8$ & $327.3$ \\

Genshin Impact
& $232.6$ & $4.65$ & $90.4$ & $210.4$ & $1@862$
& $547@134$ & $9@082$ & $2@270$ & $1.38$ & $83.4$ & $333.6$ \\

PUBG: Battlegrounds
& $197.4$ & $3.95$ & $84.6$ & $167.0$ & $1@818$
& $220@726$ & $7@233$ & $1@808$ & $2.72$ & $83.1$ & $332.5$ \\

Elden Ring Nightreign
& $168.9$ & $3.38$ & $90.0$ & $152.0$ & $1@348$
& $250@877$ & $6@630$ & $1@658$ & $2.18$ & $82.6$ & $330.2$ \\

The Witcher 3: Wild Hunt
& $158.7$ & $3.17$ & $89.4$ & $141.8$ & $1@323$
& $287@433$ & $6@290$ & $1@573$ & $1.78$ & $81.2$ & $324.6$ \\

Elden Ring
& $135.9$ & $2.72$ & $92.6$ & $125.9$ & $1@116$
& $207@706$ & $5@489$ & $1@372$ & $2.18$ & $82.6$ & $330.2$ \\

Escape from Tarkov
& $119.8$ & $2.40$ & $81.7$ & $97.9$ & $1@065$
& $129@320$ & $4@237$ & $1@059$ & $2.72$ & $83.1$ & $332.5$ \\

Apex Legends
& $116.5$ & $2.33$ & $85.3$ & $99.4$ & $1@082$
& $131@381$ & $4@305$ & $1@076$ & $2.72$ & $83.1$ & $332.5$ \\

Wuthering Waves
& $108.4$ & $2.17$ & $89.1$ & $96.5$ & $819$
& $294@088$ & $4@077$ & $1@019$ & $1.18$ & $85.2$ & $341.0$ \\

Neverness to Everness
& $101.3$ & $2.03$ & $89.3$ & $90.4$ & $800$
& $235@145$ & $3@903$ & $976$ & $1.38$ & $83.4$ & $333.6$ \\

Assassin's Creed
& $51.1$ & $1.02$ & $87.0$ & $44.5$ & $338$
& $34@401$ & $1@963$ & $491$ & $4.65$ & $81.5$ & $326.1$ \\

Black Myth: Wukong
& $51.1$ & $1.02$ & $91.3$ & $46.6$ & $414$
& $76@971$ & $2@034$ & $509$ & $2.18$ & $82.6$ & $330.2$ \\

Watch Dogs 2
& $38.5$ & $0.77$ & $85.0$ & $32.7$ & $249$
& $25@324$ & $1@445$ & $361$ & $4.65$ & $81.5$ & $326.1$ \\

Honor of Kings: World
& $35.9$ & $0.72$ & $87.3$ & $31.3$ & $277$
& $81@429$ & $1@352$ & $338$ & $1.38$ & $83.4$ & $333.6$ \\

\midrule
\textbf{Total / Avg.}
& $5@000$ & $100.00$ & $86.8$ & $4@341$ & $41@103$
& $5@947@588$ & $189@158$ & $47@290$
& $2.63$ & $82.6$ & $330.4$ \\
\bottomrule
\end{tabular}%
}
\vspace{-5.3pt}
\end{table*}

\begin{table*}[!h]
\centering
\caption{\textbf{Comparison with existing gameplay datasets.}
Large-scale denotes at least $3@000$ hours of gameplay. AAA-focused indicates that more than half of the included game titles are
AAA games. Direct human actions refer to cases where the majority of action labels
are recorded directly from human players rather than inferred, extracted, or generated.}
\vspace{-7pt}
\label{tab:dataset_comparison}
\setlength{\tabcolsep}{4pt}
\renewcommand{\arraystretch}{1.08}
\resizebox{\textwidth}{!}{%
\begin{tabular}{l*{6}{c}l}
\toprule
\multirow{2.5}{*}{\textbf{Dataset}} &
\multicolumn{2}{c}{\textbf{Data Scope}} &
\multicolumn{4}{c}{\textbf{Annotation Properties}} &
\multirow{2.5}{*}{\textbf{Notes}} \\
\cmidrule(lr){2-3}
\cmidrule(lr){4-7}
&
\shortstack{\textbf{Large}\\\textbf{Scale}} &
\shortstack{\textbf{AAA-}\\\textbf{Focused}} &
\shortstack{\textbf{Direct}\\\textbf{Human Actions}} &
\shortstack{\textbf{Instruction}\\\textbf{Annotations}} &
\shortstack{\textbf{Dense}\\\textbf{Instructions}} &
\shortstack{\textbf{Multi-}\\\textbf{Horizon}} &
\\
\midrule
\rowcolor{gray!15}
\multicolumn{8}{c}{\textit{Single-Game Datasets}} \\
\midrule
MineRL~\citep{minerl}
& \xmark & \xmark & \cmark & \xmark & \xmark & \xmark
& Minecraft; human actions; no textual instructions. \\
MineDojo~\citep{minedojo}
& \cmark & \xmark & \xmark & \xmark & \xmark & \xmark
& Web Minecraft videos; no actions or instructions. \\
VPT~\citep{vpt}
& \cmark & \xmark & \xmark & \xmark & \xmark & \xmark
& Web Minecraft videos; IDM actions; no instructions. \\
STEVE-1~\citep{steve1}
& \xmark & \xmark & \cmark & \cmark & \xmark & \xmark
& Minecraft; only 10K short-term text instructions. \\
PLAICraft~\citep{plaicraft}
& \cmark & \xmark & \cmark & \xmark & \xmark & \xmark
& Minecraft; human actions; no textual instructions. \\
WildWorld~\citep{wildworld}
& \xmark & \cmark & \xmark & \xmark & \xmark & \xmark
& Monster Hunter Wilds; AI actions; no instructions.  \\
EgoCS-400K~\citep{egocs400}
& \cmark & \cmark & \xmark & \xmark & \xmark & \xmark
& CS; replay-derived action labels; no instructions. \\

\midrule
\rowcolor{gray!15}
\multicolumn{8}{c}{\textit{Multi-Game Datasets}} \\
\midrule
NitroGen~\citep{nitrogen}
& \cmark & \xmark & \xmark & \xmark & \xmark & \xmark
& Gamepad-overlay actions without text instructions. \\
Open-P2P~\citep{openp2p}
& \cmark & \xmark & \cmark & \cmark & \xmark & \xmark
& Majority non-AAA games; sparse text instructions. \\
D2E~\citep{d2e}
& \xmark & \xmark & \xmark & \xmark & \xmark & \xmark
& Mostly IDM-inferred actions; no text instructions.  \\
%GameWorld~\citep{gameworld}
%& \xmark & \xmark & \xmark & \xmark & \xmark & \xmark
%& Browser mini-games; no action or instruction labels. \\
%GameVerse~\citep{gameverse}
%& \xmark & \xmark & \xmark & \xmark & \xmark & \xmark
%& $15$ games without action or instruction annotations. \\
Gaming-500-hours~\citep{gaming500h}
& \xmark & \xmark & \cmark & \xmark & \xmark & \xmark
& Majority non-AAA games; no textual instructions. \\
\midrule
\textbf{\data{} (Ours)}
& \cmark & \cmark & \cmark & \cmark & \cmark & \cmark
& \textbf{AAA; human actions; dense multi-horizon instructions.} \\
\bottomrule
\end{tabular}%
}
\vspace{-7pt}
\end{table*}

\noindent \textbf{Overall Workflow.} As illustrated in \reffig{}~\ref{fig:anno-pipe}, action-aware segmentation first partitions a recording of a complete gameplay session into $L_1$ clips, each annotated with a short-horizon operation. Using the $L_1$ instructions as semantic cues, we merge adjacent $L_1$ clips into $L_2$ clips and annotate them with medium-horizon goals. We then merge $L_2$ clips into $L_3$ clips and annotate them with long-horizon strategies. The \anno{} proceeds bottom-up via alternating temporal merging and instruction annotation across all temporal horizons. See the appendix for additional details regarding our prompt design, segmentation, instructions, and keybinds.

\subsection{\data{}} \label{sec:3.3}

Using our annotator, we construct \data{} as a large-scale corpus with gameplay videos, actions, and multi-horizon instructions across diverse games. We report data statistics and comparisons in this section.

\noindent \textbf{Dataset Statistics.} We present detailed statistics of our \data{} in \reftab{}~\ref{tab:data_statistics}. First, regarding data scale and proportion, the corpus comprises \lengthgamesplit{} of human gameplay across $21$ game titles, spanning diverse genres such as open-world, action role-playing, competitive shooter, sandbox survival, and creature-collecting adventure games. Valorant contributes the largest share with $617.5$ hours ($12.35\%$), while even the smallest entry, Honor of Kings: World, contains $35.9$ hours ($0.72\%$). All recordings are collected from $100$ experienced players who receive extensive training on the recording and upload procedures. We filter out low-quality or uninformative content using automated rules and VLM assessments, \textit{e.g.}, prolonged action-free cutscenes and switches away from the game window. After filtering, $4@341$ hours ($86.8\%$) are retained for instruction annotation, while the synchronized keyboard-mouse actions remain available for all \lengthgamesplit{}. Second, in terms of annotation counts, \data{} contains $411.03$ million keyboard-mouse action events, averaging $22.84$ events per second. The action stream combines mouse inputs (movements and button presses) sampled at $20$ Hz with an average of $2.84$ keyboard events per second. The retained footage is annotated with $6@184@036$ textual instructions, including $5@947@588$ short-horizon operations, $189@158$ medium-horizon goals, and $47@290$ long-horizon strategies. Third, turning to instruction spans, the average duration increases across horizons, \textit{i.e.}, $2.63$ seconds for $L_1$, $82.6$ seconds for $L_2$, and $330.4$ seconds for $L_3$. Each span measures the temporal coverage of one distinct instruction, with all frames in the corresponding clip sharing the same instruction at that horizon. Thus, every frame is simultaneously aligned with instructions at all three horizons, yielding our dense multi-horizon instruction pyramid across diverse AAA gameplay scenarios.

\noindent \textbf{Comparisons with Other Datasets.} We compare \data{} with existing datasets in
\reftab{}~\ref{tab:dataset_comparison}. Our dataset is the first
corpus that jointly provides large-scale AAA gameplay, direct
human actions, and dense multi-horizon instructions. First, for data scope, most datasets are restricted to a single game. For example, MineDojo~\citep{minedojo}, VPT~\citep{vpt}, and PLAICraft~\citep{plaicraft} cover only Minecraft. WildWorld~\citep{wildworld} and EgoCS-400K~\citep{egocs400} are confined to Monster Hunter Wilds and Counter-Strike. The few multi-game
datasets span predominantly non-AAA games, \textit{e.g.},
NitroGen~\citep{nitrogen} and Open-P2P~\citep{openp2p}. In contrast, \data{} contains $5@000$ hours of gameplay across $21$ titles, with a primary focus on AAA games. Second, for action annotations, several datasets lack directly recorded human controls. For example, MineDojo~\citep{minedojo} provides web Minecraft videos without action annotations, while some other datasets obtain action labels using IDMs~\citep{vpt,d2e}, gamepad segmentation~\citep{nitrogen}, in-game AI~\citep{wildworld}, or game replays~\citep{egocs400}. By comparison, \data{} captures synchronized keyboard-mouse actions from $100$ experienced human players across the entire dataset. Third, most existing datasets omit instructions, \textit{e.g.}, VPT~\citep{vpt}, NitroGen~\citep{nitrogen}, D2E~\citep{d2e}, and
gaming-500-hours~\citep{gaming500h}. Among the few exceptions, STEVE-1~\citep{steve1} contains $10$K short-term instructions, while Open-P2P~\citep{openp2p} provides one instruction every few
minutes. Our dataset contains a total of $6@184@036$ distinct multi-horizon instructions with dense per-frame coverage.

\begin{figure}[!t]
    \centering
    \vspace{-1pt}  
    \includegraphics[width=\textwidth,trim=17 0 11 0,clip]{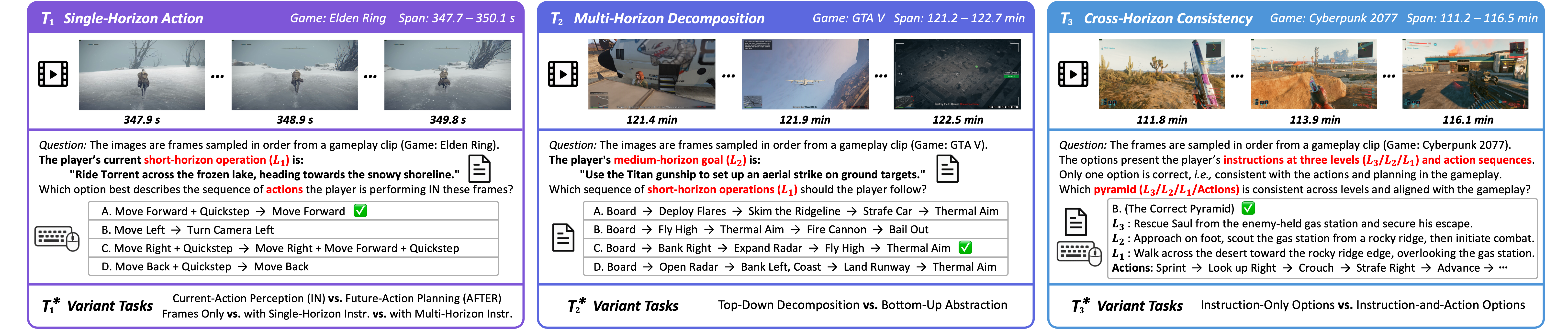}
    \vspace{-14pt}
    \caption{\textbf{The offline track of \bench{}.} It comprises three primary tasks and ten variant tasks. Given sampled frames and the short-horizon instruction $L_1$, single-horizon action $T_1$ requires models to determine the correct action sequence. Multi-horizon decomposition $T_2$ evaluates whether models can decompose a medium-horizon goal $L_2$ into the correct sequence of short-horizon operations $L_1$. Cross-horizon consistency $T_3$ assesses the overall consistency across frames, $L_1$–$L_3$ instructions, and actions. Variant tasks $T^*$ provide additional diagnostics and insights into different gameplay abilities. Due to space constraints, $L_1$ instructions are abridged in $T_2$. Only the correct option appears in $T_3$.} 

  \label{fig:offline-demo} 
  \vspace{-7pt}
\end{figure}

\subsection{\bench{}} \label{sec:3.4}

Leveraging our data, we develop \bench{}, a unified benchmark that evaluates multi-horizon gameplay capabilities across diverse model families through the reproducible offline and stepwise online tracks. 
%The offline track enables reproducible evaluation, while the online track tests long-horizon gameplay stepwise.

%The offline track provides standardized and reproducible evaluation, while the online track assesses long-horizon gameplay through a stepwise protocol.

\noindent \textbf{Reproducible Offline Track.} The offline track formulates gameplay evaluation as standardized MCQs derived from the aligned frames, multi-horizon instructions, and player actions in \data{}. It contains thousands of questions organized into three primary tasks and ten variant tasks. High-quality annotations, standardized inputs and options, as well as independence from game environments and agent harnesses enable unified and reproducible comparisons across the VLMs, UMMs, GUI agents, coding agents, and game agents. 

To be specific, as shown in \reffig{}~\ref{fig:offline-demo}, the single-horizon action task ($T_1$) asks models to make the correct action decision based on input frames and the short-horizon instruction ($L_1$). The correct option is constructed by mapping ground-truth keyboard-mouse controls to textual action sequences using game-specific keybinds, which enables unified evaluation across games and models without being constrained by heterogeneous action spaces. $T_1$ provides a rigorous test centered on action perception. Next, the multi-horizon decomposition task ($T_2$) requires models to identify the ordered sequence of short-horizon operations for realizing a medium-horizon goal ($L_2$). The correct option presents the $L_1$ sequence corresponding to the target $L_2$ goal within the pyramid, whereas the distractors differ in their composition and temporal order. $T_2$ evaluates top-down goal decomposition and temporal planning from abstract goals to concrete operations. Finally, the cross-horizon consistency task ($T_3$) assesses the overall consistency across videos, instructions, and actions. The correct option consists of the aligned $L_3$, $L_2$, and $L_1$ instructions, along with the action sequence from the pyramid. $T_3$ focuses on integrated understanding across visual observations, textual instructions, and gameplay actions.

Moreover, a series of variant tasks ($T^*$) can provide additional diagnostics and analyses of different gameplay capabilities. For $T_1^*$, we compare current-action perception and future-action planning under three input settings, including frames only, frames with a short-horizon instruction, and frames with multi-horizon instructions.  The comparison reveals how textual instructions influence action decisions and planning. For $T_2^*$, we compare top-down decomposition from $L_2$ goals to $L_1$ operations with bottom-up abstraction in the reverse direction. The comparison clarifies the rationale for our bottom-up annotation pipeline and top-down benchmark design. As for $T_3^*$, we compare instruction-only versus instruction-and-action options. The comparison helps identify whether the performance bottlenecks arise from goal planning or action decoding.

\begin{figure}[!t]
    \centering
    \vspace{-1pt}  
    \includegraphics[width=\textwidth,trim=0 0 2 0,clip]{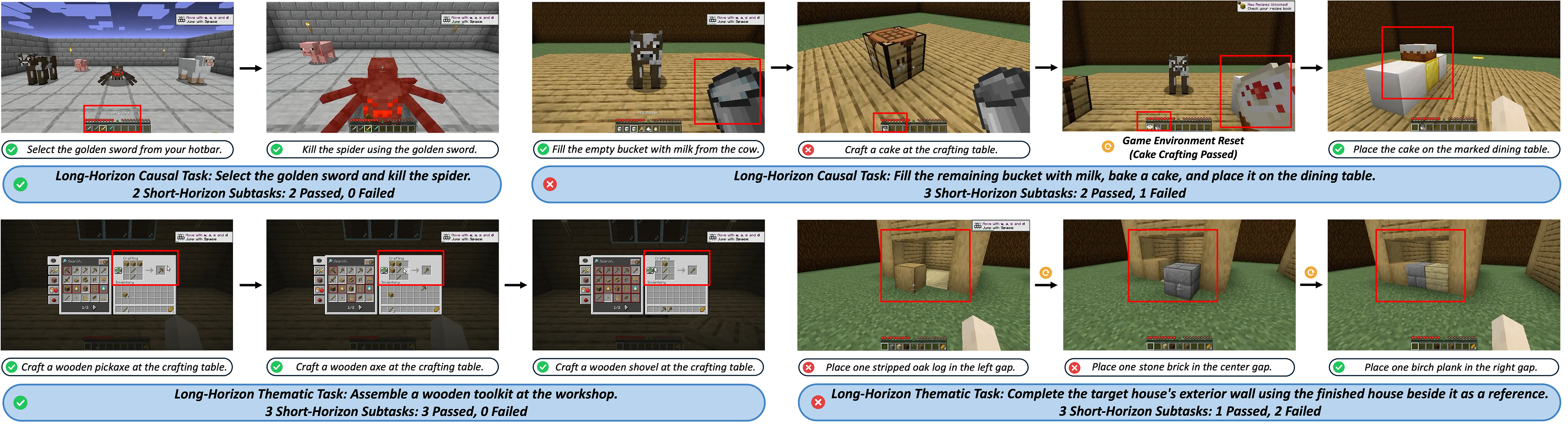}
    \vspace{-14pt}
    \caption{\textbf{The online track of \bench{}.} It contains long-horizon causal and thematic tasks, each comprising multiple verifiable short-horizon subtasks. The causal tasks can only be completed in a prescribed order because of dependencies between successive subtasks. The thematic tasks contain subtasks that share a common theme but can be performed in any order. The figure shows results from Gemini 3.6 Flash~\citep{gemini36flash}, with arrows indicating its actual execution order. After a subtask fails, the environment is reset to the corresponding success state so that evaluation can continue. A long-horizon task is considered passed only when all constituent subtasks succeed.} 

  \label{fig:online-demo} 
  \vspace{-7pt}
\end{figure}

\noindent \textbf{Stepwise Online Track.} As a complementary evaluation, our online track tests whether offline scores are positively associated with gameplay capabilities. It evaluates long-horizon gameplay through order-dependent causal tasks and order-flexible thematic tasks, each comprising $2$--$6$ verifiable short-horizon subtasks. When a subtask fails, the game is reset to the corresponding success state to prevent the failure from affecting subsequent steps. Beyond the aggregate success rates in previous work~\citep{lumine,gametars,sima2}, our stepwise protocol can localize failures to specific stages within long-horizon objectives.

Specifically, the online track comprises $10$ causal and $10$ thematic tasks with $62$ short-horizon subtasks. Due to the inaccessibility of underlying game states in most AAA titles, we implement the online track in Minecraft, as in prior work~\citep{mineexp,mcu,jarvisvla,minestudio,minedojo}. Four example tasks are presented in \reffig{}~\ref{fig:online-demo} with the actual execution order and results of Gemini 3.6 Flash~\citep{gemini36flash}. The steps within each causal task have sequential dependencies, e.g., first collecting milk, then crafting a cake, and finally placing it on the dining table. For thematic tasks, the subtasks share a common theme and can be completed in any order, e.g., crafting a wooden pickaxe, axe, and shovel for a toolkit. In the failure cases, Gemini 3.6 Flash understands intended operations yet still commits execution errors, \textit{e.g.}, placing objects in inaccurate positions or failing to follow the correct recipe when making a cake.

\noindent \textbf{Comparisons with Other Benchmarks.} We compare gameplay benchmarks in
\reftab{}~\ref{tab:benchmark_comparison}. First, with a primary focus on AAA titles, \bench{} evaluates \nummodel{} models on $21$ games, \textit{e.g.}, VLMs, UMMs, GUI, coding, and game agents. Prior work tests limited model types mainly on Minecraft~\citep{minedojo,mcu,mineexp} or non-AAA games~\citep{balrog,gameverse,gameworld}. Second, multi-horizon instructions and human actions enable reliable evaluation across nested horizons. Third, beyond online success rates~\citep{minedojo,gameworld,lumine} or progress scores~\citep{balrog,mineexp}, we provide the reproducible offline track with thousands of MCQs and the stepwise online track with failure localization. Refer to the appendix for more details about our \bench{}.

\begin{table*}[!t]
\centering
\caption{\textbf{Comparison with existing gameplay benchmarks.} Model diversity denotes evaluation across at least three model paradigms, \textit{e.g.}, VLMs, UMMs, GUI, coding, and game agents. Our \bench{} tests \nummodel{} models of five types.}
\vspace{-7pt}
\label{tab:benchmark_comparison}
\setlength{\tabcolsep}{5pt}
\renewcommand{\arraystretch}{1.1}
\resizebox{\textwidth}{!}{%
\begin{tabular}{l*{6}{c}l}
\toprule
\multirow{2.5}{*}{\textbf{Benchmark}} &
\multicolumn{2}{c}{\textbf{Benchmark Scope}} &
\multicolumn{2}{c}{\textbf{Annotation Properties}} &
\multicolumn{2}{c}{\textbf{Evaluation Settings}} &
\multirow{2.5}{*}{\textbf{Notes}} \\
\cmidrule(lr){2-3}\cmidrule(lr){4-5}\cmidrule(lr){6-7}
&
\shortstack{\textbf{Model}\\\textbf{Diversity}} &
\shortstack{\textbf{AAA-}\\\textbf{Focused}} &
\shortstack{\textbf{Direct Human}\\\textbf{Actions}} &
\shortstack{\textbf{Multi-}\\\textbf{Horizon}} &
\shortstack{\textbf{Reproducible}\\\textbf{Offline}} &
\shortstack{\textbf{Stepwise}\\\textbf{Online}} &
\\
\midrule
\rowcolor{gray!15}
\multicolumn{8}{c}{\textit{Single-Game Benchmarks}} \\
\midrule
MineDojo~\citep{minedojo}
& \xmark & \xmark & \xmark & \xmark & \xmark & \xmark
& Online success rates in Minecraft; the game agents only. \\
MCU~\citep{mcu}
& \xmark & \xmark & \xmark & \xmark & \xmark & \xmark
& Minecraft; VLM-based online scores; game agents only. \\
MineExplorer~\citep{mineexp}
& \xmark & \xmark & \xmark & \xmark & \xmark & \cmark
& Online milestone rates in Minecraft; the VLMs only. \\

%MineStudio~\citep{minestudio}
%& \xmark & \xmark & \xmark & \xmark & \xmark & \xmark
%& Minecraft; configurable tasks with video-based evaluation. \\

\midrule
\rowcolor{gray!15}
\multicolumn{8}{c}{\textit{Multi-Game Benchmarks}} \\
\midrule
BALROG~\citep{balrog}
& \xmark & \xmark & \xmark & \xmark & \xmark & \cmark
& Non-AAA; the online progress scores; LLMs and VLMs. \\
VideoGameBench~\citep{videogamebench}
& \xmark & \xmark & \xmark & \xmark & \xmark & \xmark
& Retro and classic games; online scores and VLMs only. \\
Orak~\citep{orak}
& \xmark & \xmark & \xmark & \xmark & \xmark & \xmark
& Majority non-AAA; the online scores; LLMs and VLMs. \\
GameWorld~\citep{gameworld}
& \xmark & \xmark & \xmark & \xmark & \xmark & \xmark
& Browser mini-games only; online scores; VLM agents. \\
GameVerse~\citep{gameverse}
& \xmark & \xmark & \xmark & \xmark & \xmark & \cmark
& Mostly non-AAA; online milestone scores; VLM agents. \\

\midrule
\textbf{\bench{} (Ours)}
& \cmark & \cmark & \cmark & \cmark & \cmark & \cmark
& \textbf{AAA; reproducible offline; stepwise online; diverse models.} \\
\bottomrule
\end{tabular}%
}
\vspace{-7pt}
\end{table*}

\section{Experiments}

We conduct extensive experiments and systematic evaluations using our \bench{}. We describe the evaluation protocols and implementation details in \refsec{}~\ref{sec:4.1}. The testing results on our offline primary tasks and variant tasks are presented in \refsec{}~\ref{sec:4.2} and \refsec{}~\ref{sec:4.3}, respectively. Besides, \refsec{}~\ref{sec:4.4} reports the results from our stepwise online track, which examines the relations between offline scores and online gameplay abilities.

\subsection{Evaluation Protocols and Details}\label{sec:4.1}

\noindent \textbf{Offline Track.} We evaluate \nummodel{} models spanning five categories, including $35$ general-purpose VLMs, $4$ UMMs, $2$ GUI agents, $3$ coding agents, and $3$ dedicated game agents. The $44$ models of the first four types natively support question answering. We adopt unified MCQs for all offline tasks. Each model receives the same four-option questions and directly outputs the selections, ensuring standardized and reproducible evaluation. 

On the other hand, the dedicated game agents~\citep{openp2p,nitrogen,jarvisvla} are trained to produce action sequences without supporting question answering or reasoning. Thus, we evaluate these models only on the single-horizon action task ($T_1$) by matching their action predictions against the four options via sequence alignment~\citep{needleman1970general}. The option with the highest alignment score is taken as the selected answer. If all four scores fall below a threshold, indicating no meaningful overlap with any option, the prediction is marked incorrect, \textit{e.g.}, an empty action sequence. For NitroGen~\citep{nitrogen}, we additionally map its gamepad actions to the keyboard-and-mouse trajectories using our keybinds. 

As for the implementation details, each of the three primary tasks ($T_1$--$T_3$) contains $1@000$ MCQs, while each of the ten variants contains $200$ MCQs. In total, the offline track consists of $5@000$ MCQs for high-confidence and reliable evaluation. Some models are evaluated through their official APIs, while the others are served locally on GPUs. The inference of dedicated game agents follows their official checkpoints, rollout procedures, and context-window sizes, whereas the other four model categories are given the same number of input frames. We report MCQ accuracy for all models and divide them into four performance tiers based on their rankings.

\noindent \textbf{Online Track.} Due to evaluation costs, we sample three models per offline tier, yielding $12$ models for online testing. We construct ten causal and ten thematic tasks with $62$ short-horizon subtasks. The thematic tasks span diverse themes, \textit{e.g.}, resource gathering, crafting, farming, building, and combat. A multi-step long-horizon task is considered passed only when the model completes it end-to-end without intermediate failures and game resets. Besides, each subtask is allocated a maximum budget of $400$ model calls. Any subtask not completed within this budget is marked as failed. The task outcomes are automatically determined by the underlying game states. We report success rates for both short-horizon subtasks and long-horizon tasks.

\subsection{Offline Primary Tasks}\label{sec:4.2}

\noindent\textbf{Benchmark Discriminability.} We first test the $44$ models with question-answering abilities in \reftab{}~\ref{tab:offline_primary}. The mean accuracy across our benchmark is $64.7\%$, substantially above the $25\%$ random baseline, suggesting that the designed tasks are solvable yet challenging for meaningful evaluation. Among the evaluated models, the accuracy ranges from $44.6\%$ to $80.2\%$, spanning $35.6$ percentage points and clearly distinguishing models in different capability tiers. At the task level, the results reveal a coherent difficulty hierarchy, \textit{i.e.}, $57.3\%$ on $T_1$, $65.1\%$ on $T_2$, and $71.6\%$ on $T_3$. $T_1$ is the most difficult task because it requires fine-grained action prediction under strict correctness criteria from keyboard-mouse trajectories. $T_3$ emphasizes cross-horizon understanding and exhibits the highest accuracy because the options provide rich information from multi-horizon instructions and actions. $T_2$ lies between $T_1$ and $T_3$, focusing on the decomposition and planning of complex goals. Overall, our \bench{} demonstrates an appropriate task-difficulty gradient and clear model discriminability.

\noindent\textbf{General-Purpose VLMs.} Proprietary models dominate the top of the leaderboard, occupying ten of the eleven positions in Tier~1. GPT-6-Astra ranks first with $80.2\%$ overall accuracy, followed by Gemini 3.8 Flash and Gemini 3.7 Flash with $77.3\%$ and $76.7\%$, respectively. Kimi-K3 is the only open-weight model in Tier~1, ranking sixth with $74.5\%$. Recent open-weight models also show competitive performance, \textit{e.g.}, Qwen3.8-27B, Kimi-K2.6, and Gemma 4 31B-IT rank 16th, 17th, and 20th, surpassing several proprietary models. Within the same model family, newer or larger variants generally perform better. For example, overall accuracy increases from $62.8\%$ for GPT-5.2 to $73.2\%$ for GPT-5.5, $74.8\%$ for GPT-5.6 Sol, and $80.2\%$ for GPT-6-Astra. Similar trends are also observed in other families, such as Gemini, Kimi, Qwen, Doubao, InternVL, and GLM.

\begin{table*}[!t]
\centering
\caption{\textbf{Offline results of the primary tasks $\boldsymbol{T_1}$--$\boldsymbol{T_3}$.}
The table reports results for general VLMs, UMMs, GUI agents, and coding agents with question-answering abilities. Accuracies are reported as percentages. Overall denotes the mean accuracy across three tasks. Models are ordered from higher to lower performance and divided into four tiers. The best and second-best
metrics are marked in bold and underlined. The final row reports the average accuracy for all models.}
\vspace{-7pt}
\label{tab:offline_primary}
\setlength{\tabcolsep}{5pt}
\renewcommand{\arraystretch}{1.1}
\resizebox{\textwidth}{!}{%
\begin{tabular}{llccccllcccc}
\toprule
\rowcolor{gray!15}
\multicolumn{6}{c}{\textbf{Tier 1}} &
\multicolumn{6}{c}{\textbf{Tier 2}} \\
\midrule
\textbf{Rank} & \textbf{Model} &
$\boldsymbol{T_1}$ & $\boldsymbol{T_2}$ & $\boldsymbol{T_3}$ & \textbf{Overall} &
\textbf{Rank} & \textbf{Model} &
$\boldsymbol{T_1}$ & $\boldsymbol{T_2}$ & $\boldsymbol{T_3}$ & \textbf{Overall} \\
\midrule

\medal{gold}{1}   & GPT-6-Astra           & \textbf{69.4} & 79.6 & \textbf{91.5} & \textbf{80.2} &
\textbf{12} & Doubao-Seed-2.0-Pro         & 59.2 & 74.9 & 80.0 & 71.4 \\

\medal{silver}{2} & Gemini 3.8 Flash      & 65.9 & \textbf{81.2} & \underline{84.8} & \underline{77.3} &
\textbf{13} & Claude Fable 5              & 62.9 & 72.7 & 78.1 & 71.2 \\

\medal{bronze}{3} & Gemini 3.7 Flash      & \underline{66.2} & \underline{80.1} & 83.9 & 76.7 &
\textbf{14} & GPT-5.6 Terra               & 62.1 & 75.6 & 74.4 & 70.7 \\

\textbf{4}  & Gemini 3.6 Flash            & 62.3 & 79.1 & 84.6 & 75.3 &
\textbf{15} & Qwen3.7-Plus                & 58.8 & 75.9 & 71.5 & 68.7 \\

\textbf{5}  & GPT-5.6 Sol                 & 65.3 & 77.5 & 81.5 & 74.8 &
\textbf{16} & Qwen3.8-27B                 & 59.9 & 77.4 & 68.1 & 68.5 \\

\textbf{6}  & Kimi-K3                     & 64.4 & 79.3 & 79.7 & 74.5 &
\textbf{17} & GPT-5.6 Luna                & 61.6 & 70.7 & 71.1 & 67.8 \\

\textbf{7}  & Gemini 3.5 Flash            & 64.9 & 77.4 & 80.8 & 74.4 &
\textbf{17} & Kimi-K2.6                   & 61.2 & 73.2 & 69.0 & 67.8 \\

\textbf{8}  & GPT-5.5                     & 64.6 & 78.9 & 76.2 & 73.2 &
\textbf{19} & Doubao-Seed-2.0-Lite        & 55.2 & 69.0 & 76.3 & 66.8 \\

\textbf{9}  & Gemini 3.1 Pro              & 64.5 & 75.8 & 78.2 & 72.8 &
\textbf{20} & Gemma 4 31B-IT              & 53.8 & 68.6 & 76.6 & 66.3 \\

\textbf{10} & Doubao-Seed-2.1-Turbo       & 63.1 & 79.4 & 75.5 & 72.7 &
\textbf{21} & MiniMax-M3                  & 57.8 & 68.5 & 70.6 & 65.6 \\

\textbf{11} & Doubao-Seed-2.1-Pro         & 62.9 & 76.7 & 76.6 & 72.1 &
\textbf{22} & Qwen3-VL-235B-A22B-Thinking & 58.2 & 71.6 & 65.9 & 65.2 \\

\midrule
\rowcolor{gray!15}
\multicolumn{6}{c}{\textbf{Tier 3}} &
\multicolumn{6}{c}{\textbf{Tier 4}} \\
\midrule
\textbf{Rank} & \textbf{Model} &
$\boldsymbol{T_1}$ & $\boldsymbol{T_2}$ & $\boldsymbol{T_3}$ & \textbf{Overall} &
\textbf{Rank} & \textbf{Model} &
$\boldsymbol{T_1}$ & $\boldsymbol{T_2}$ & $\boldsymbol{T_3}$ & \textbf{Overall} \\
\midrule

\textbf{23} & Claude Opus 4.8             & 59.1 & 62.1 & 73.1 & 64.8 &
\textbf{34} & Qwen3.6-35B-A3B             & 53.3 & 52.9 & 66.3 & 57.5 \\

\textbf{24} & Claude Sonnet 5             & 54.4 & 57.7 & 82.0 & 64.7 &
\textbf{35} & Qwen3-Omni-30B-A3B-Instruct & 52.5 & 55.6 & 63.7 & 57.3 \\

\textbf{25} & Step-3.7-Flash              & 57.5 & 71.3 & 63.4 & 64.1 &
\textbf{36} & GELab-Zero-4B-Preview       & 51.8 & 46.8 & 72.7 & 57.1 \\

\textbf{26} & Qwen3-VL-235B-A22B-Instruct & 55.1 & 62.1 & 73.1 & 63.4 &
\textbf{37} & Qwen2.5-VL-7B-Instruct      & 52.0 & 54.3 & 63.7 & 56.7 \\

\textbf{27} & GLM-5V-Turbo                & 53.4 & 65.4 & 70.8 & 63.2 &
\textbf{38} & InternVL3.5-8B              & 51.8 & 50.2 & 65.5 & 55.8 \\

\textbf{28} & GPT-5.2                     & 58.9 & 58.9 & 70.6 & 62.8 &
\textbf{39} & GLM-4.1V-9B-Thinking        & 54.9 & 54.8 & 55.3 & 55.0 \\

\textbf{29} & Qwen3.5-397B-A17B           & 53.3 & 60.1 & 73.5 & 62.3 &
\textbf{40} & GPT-4o                      & 55.7 & 46.3 & 61.3 & 54.4 \\

\textbf{30} & BAGEL-7B-MoT                & 51.7 & 57.3 & 75.1 & 61.4 &
\textbf{41} & UI-TARS-1.5-7B              & 49.1 & 47.1 & 62.9 & 53.0 \\

\textbf{31} & Qwen2.5-VL-32B-Instruct     & 54.8 & 58.0 & 69.7 & 60.8 &
\textbf{42} & Ovis-U1-3B                  & 44.7 & 39.6 & 59.2 & 47.8 \\

\textbf{32} & Step3-VL-10B                & 55.4 & 64.7 & 61.7 & 60.6 &
\textbf{43} & InternVL3.5-2B              & 45.1 & 40.3 & 52.7 & 46.0 \\

\textbf{33} & SenseNova-U1-8B-MoT         & 48.2 & 59.2 & 67.5 & 58.3 &
\textbf{44} & InternVL-U-4B               & 46.1 & 37.9 & 49.8 & 44.6 \\

\midrule
\rowcolor{gray!15}
\multicolumn{12}{c}{
\textbf{Average over all models:}\quad
$\boldsymbol{T_1=57.3}$ \quad
$\boldsymbol{T_2=65.1}$ \quad
$\boldsymbol{T_3=71.6}$ \quad
\textbf{Overall} $\boldsymbol{= 64.7}$
} \\
\bottomrule
\end{tabular}%
}
\vspace{-7pt}
\end{table*}

\noindent\textbf{Unified Multimodal Models.} We evaluate four UMMs, including BAGEL-7B-MoT~\citep{bagel}, SenseNova-U1-8B-MoT~\citep{sensenova}, Ovis-U1-3B~\citep{ovis}, and InternVL-U-4B~\citep{internvl-u}. All four models fall into Tier~3 or Tier~4, indicating that current UMMs remain less competitive in gameplay tasks. The larger models perform better, \textit{i.e.}, BAGEL-7B-MoT achieves $61.4\%$ and ranks 30th, while SenseNova-U1-8B-MoT achieves $58.3\%$ and ranks 33rd. In comparison, Ovis-U1-3B and InternVL-U-4B achieve only $47.8\%$ and $44.6\%$. To examine potential interference between understanding and generation, we compare InternVL-U-4B with its understanding backbone, InternVL3.5-2B. Despite having additional generation parameters, InternVL-U-4B underperforms InternVL3.5-2B by $1.4$ percentage points. The result suggests that unifying understanding and generation could introduce interference, offsetting gains from increased capacity.

\noindent\textbf{GUI Agents.} We evaluate two GUI agents, GELab-Zero-4B-Preview~\citep{stepgui} and UI-TARS-1.5-7B~\citep{uitars1}, on our benchmark. Both models fall into Tier~4, indicating limited transfer from computer-use environments to AAA gameplay. GELab-Zero-4B-Preview achieves $57.1\%$ overall accuracy and ranks 36th, outperforming UI-TARS-1.5-7B by $4.1$ percentage points. UI-TARS-1.5-7B achieves $53.0\%$ and ranks 41st, trailing its general-purpose baseline, Qwen2.5-VL-7B-Instruct, by $3.7$ percentage points. These results suggest that domain-specific post-training can weaken the model generalization across domains.

{\parfillskip=0pt
\noindent\textbf{Coding Agents.} We evaluate three coding agents, including Claude Sonnet 5, Opus 4.8, and Fable 5. Among the three models, Claude Fable 5 performs best, ranking 13th in Tier~2 with $71.2\%$ overall accuracy. Claude Opus 4.8 and Sonnet 5 fall into Tier~3, ranking 23rd and 24th with $64.8\%$ and $64.7\%$, respectively. Despite the focus on coding abilities, these models generalize well to gameplay and achieve competitive performance on our benchmark. Their advantage over the GUI agents~\citep{uitars1,stepgui} may stem from the broader training data, stronger foundation models, and better reasoning capabilities.\par
}

\begin{table*}[!t]
\centering
\caption{\textbf{Offline results of game agents.} We evaluate NitroGen~\citep{nitrogen}, Open-P2P~\citep{openp2p}, and JARVIS-VLA~\citep{jarvisvla}, with Gemma 4 31B-IT as a reference. Game agents are trained
to predict actions without question answering or reasoning abilities. We test them only on the single-horizon action task $T_1$. In-domain denotes fine-tuning on the dataset, \textit{e.g.}, JARVIS-VLA on VPT~\citep{vpt}. Zero-shot indicates the opposite.}
\vspace{-5pt}
\label{tab:game_agent}
\setlength{\tabcolsep}{3pt}
\renewcommand{\arraystretch}{1.1}
\resizebox{0.87\textwidth}{!}{%
\begin{minipage}{0.96\textwidth}
\centering

\begin{subtable}[t]{0.49\linewidth}
\centering
\small
\begin{tabular*}{\linewidth}{
@{\extracolsep{\fill}}lccc@{}
}
\toprule
\textbf{Model} &
$\boldsymbol{T_1}$ &
\textbf{In-Domain} &
\textbf{Zero-Shot} \\
\midrule
NitroGen    & 22.0 & \xmark & \cmark \\
Open-P2P    & \underline{24.4} & \xmark & \cmark \\
JARVIS-VLA  & \textbf{28.1} & \xmark & \cmark \\
\bottomrule
\end{tabular*}
\vspace{+2pt}
\caption{\textbf{\bench{} (AAA games)}}
\label{tab:game_agent_GH}
\end{subtable}
\hfill
\begin{subtable}[t]{0.49\linewidth}
\centering
\small
\begin{tabular*}{\linewidth}{
@{\extracolsep{\fill}}lccc@{}
}
\toprule
\textbf{Model} &
$\boldsymbol{T_1}$ &
\textbf{In-Domain} &
\textbf{Zero-Shot} \\
\midrule
Open-P2P        & 26.8 & \xmark & \cmark \\
Gemma 4 31B-IT  & \underline{54.1} & \xmark & \cmark \\
JARVIS-VLA      & \textbf{54.2} & \cmark & \xmark \\
\bottomrule
\end{tabular*}
\vspace{+2pt}
\caption{\textbf{VPT (Minecraft)}}
\label{tab:game_agent_vpt}
\end{subtable}

\end{minipage}
}
\vspace{-7pt}
\end{table*}

\begin{table*}[!t]
\centering
\caption{\textbf{Effects of thinking on offline results.}
$\Delta$ denotes the change in overall accuracy. For lightweight models,
thinking often induces hallucinations, leading to performance degradation.
More capable models can leverage thinking to plan and decompose complex goals in gameplay,
thus improving the accuracy, \textit{e.g.}, GLM-5V-Turbo and
Doubao-Seed-2.1-Pro.}
\vspace{-5pt}
\label{tab:thinking_effect}
\setlength{\tabcolsep}{1.5pt}
\renewcommand{\arraystretch}{1.1}

\resizebox{0.87\textwidth}{!}{%
\begin{minipage}{\textwidth}
\centering

\begin{subtable}[t]{0.487\linewidth}
\centering
{\small
\begin{tabular*}{\linewidth}{
@{\extracolsep{\fill}}lcccccc@{}
}
\toprule
\textbf{Model} &
\textbf{Thinking} &
$\boldsymbol{T_1}$ &
$\boldsymbol{T_2}$ &
$\boldsymbol{T_3}$ &
\textbf{Overall} &
$\boldsymbol{\Delta}$ \\
\midrule

\multirow{2}{*}{UI-TARS-1.5-7B}
& \xmark & 49.1 & 47.1 & 62.9 & 53.0 & -- \\
& \cmark & 45.2 & 45.4 & 43.9 & 44.8
& \textcolor{red}{\textbf{-8.2}} \\

\midrule

\multirow{2}{*}{Qwen2.5-VL-7B}
& \xmark & 52.0 & 54.3 & 63.7 & 56.7 & -- \\
& \cmark & 44.4 & 52.2 & 55.7 & 50.8
& \textcolor{red}{\textbf{-5.9}} \\

\bottomrule
\end{tabular*}
}
\vspace{2pt}
\caption{\textbf{Degradation with Thinking}}
\label{tab:thinking_degrades}
\end{subtable}
\hfill
\begin{subtable}[t]{0.487\linewidth}
\centering
{\small
\begin{tabular*}{\linewidth}{
@{\extracolsep{\fill}}lcccccc@{}
}
\toprule
\textbf{Model} &
\textbf{Thinking} &
$\boldsymbol{T_1}$ &
$\boldsymbol{T_2}$ &
$\boldsymbol{T_3}$ &
\textbf{Overall} &
$\boldsymbol{\Delta}$ \\
\midrule

\multirow{2}{*}{GLM-5V-Turbo}
& \xmark & 53.2 & 56.4 & 63.1 & 57.6 & -- \\
& \cmark & 53.4 & 65.4 & 70.8 & 63.2
& \textcolor{green!60!black}{\textbf{+5.6}} \\

\midrule

\multirow{2}{*}{Doubao-Seed-2.1-Pro}
& \xmark & 56.4 & 56.8 & 75.2 & 62.8 & -- \\
& \cmark & 62.9 & 76.7 & 76.6 & 72.1
& \textcolor{green!60!black}{\textbf{+9.3}} \\

\bottomrule
\end{tabular*}
}
\vspace{2pt}
\caption{\textbf{Improvement with Thinking}}
\label{tab:thinking_improves}
\end{subtable}

\end{minipage}%
}

\vspace{-7pt}
\end{table*}

\noindent\textbf{Dedicated Game Agents.} As shown in \reftab{}~\ref{tab:game_agent}, the three game agents perform near the $25\%$ random baseline on unseen AAA games. Their scores can fall below random because action outputs that fail to match any option are counted as incorrect, \textit{e.g.}, empty action sequences. JARVIS-VLA~\citep{jarvisvla} shows strong domain dependence. After fine-tuning on VPT~\citep{vpt}, it achieves $54.2\%$ on the in-domain Minecraft data, comparable to the zero-shot Gemma 4 31B-IT at $54.1\%$. However, the accuracy drops to $28.1\%$ on unseen AAA games. NitroGen~\citep{nitrogen} and Open-P2P~\citep{openp2p} show similar limitations, as neither is trained on AAA gameplay data with aligned multi-horizon instructions and actions. These results further highlight the value of our dataset covering diverse AAA game titles and multiple temporal horizons.

\noindent\textbf{Effects of Thinking.} As shown in \reftab{}~\ref{tab:thinking_effect}, thinking does not uniformly improve offline performance. The effects vary with model reasoning quality. Enabling thinking improves the accuracy of GLM-5V-Turbo and Doubao-Seed-2.1-Pro by $5.6$ and $9.3$ percentage points. The gains are particularly pronounced on $T_2$, reaching $9.0$ and $19.9$ percentage points, suggesting that capable models can leverage thinking to decompose and plan complex goals. In contrast, thinking reduces the accuracy of UI-TARS-1.5-7B and Qwen2.5-VL-7B by $8.2$ and $5.9$ percentage points. Qualitative inspection indicates that lightweight models can produce hallucinated reasoning, thereby misleading action prediction rather than improving it. The results indicate that the efficacy of thinking in gameplay depends on whether it enables models to decompose and plan complex goals correctly. The findings reveal an inherent trade-off between reasoning ability and model efficiency in real-time gameplay.

\subsection{Offline Variant Tasks}\label{sec:4.3}

Given the evaluation costs, we test four models on $200$ randomly sampled questions for each of the ten variants.

\noindent \textbf{Action Perception and Planning.} As shown in \reftab{}~\ref{tab:t1_four_model_ablation}, all four models consistently perform better on current-action perception than future-action planning across the three input settings. On average, perception outperforms planning by $9.9$ percentage points with frames only, $23.8$ points with short-horizon instructions, and $18.2$ points with multi-horizon instructions. The consistent gap indicates that models more reliably recognize actions depicted in the observed frames than determine appropriate future actions. Future-action planning remains more challenging because it requires inferring subsequent behavior from temporal context. Nevertheless, the capability of planning is essential to gameplay, especially for the complex long-horizon tasks.

\noindent \textbf{Visual Inputs and Textual Instructions.} \reftab{}~\ref{tab:t1_four_model_ablation} shows that adding short-horizon instructions to the visual inputs consistently improves action perception across all four models. Average accuracy increases by $14.1$ percentage points, from $44.5\%$ with frames only to $58.6\%$ with short-horizon instructions. Notably, the improvement is achieved solely by adding the instructions to the model inputs, without any fine-tuning on our annotations. The results indicate that visual observations alone~\citep{nitrogen} may not fully specify task intent, whereas textual instructions can help models to identify the correct operations for the specified game goals. The finding underscores the value of our instructions as inputs for model training and evaluation in gameplay.

\begin{table*}[!t]
\centering

\caption{\textbf{Results of the offline variant tasks $\boldsymbol{T_1^*}$.} The table presents six variants of the primary task $T_1$, including current-action perception and future-action planning under three input settings, \textit{i.e.}, frames only, frames with short-horizon instructions, and frames with multi-horizon instructions. Considering the evaluation costs, we test four models using $200$ questions for each variant. Values outside parentheses denote accuracy, while parenthesized values denote changes in percentage points relative to the preceding column. The final row reports the average results across all four models.}
\vspace{-7pt}
\label{tab:t1_four_model_ablation}
\setlength{\tabcolsep}{4pt}
\renewcommand{\arraystretch}{1.1}
\resizebox{0.87\textwidth}{!}{%
\begin{tabular}{lcccccc}
\toprule
\multirow{2}{*}{\textbf{Model}} &
\multicolumn{3}{c}{\textbf{Current-Action Perception}} &
\multicolumn{3}{c}{\textbf{Future-Action Planning}} \\
\cmidrule(lr){2-4}\cmidrule(lr){5-7}
&
\textbf{Frames Only} &
\textbf{+ Short} &
\textbf{+ Multi-Horizon} &
\textbf{Frames Only} &
\textbf{+ Short} &
\textbf{+ Multi-Horizon} \\
\midrule

Qwen2.5-VL-32B-Instruct
& 39.0
& 53.5 ($+14.5$)
& 54.5 ($+1.0$)
& 33.5
& 29.5 ($-4.0$)
& 33.5 ($+4.0$) \\

Qwen3-VL-235B-A22B-Instruct
& 41.0
& 52.0 ($+11.0$)
& 54.5 ($+2.5$)
& 37.0
& 34.0 ($-3.0$)
& 39.5 ($+5.5$) \\

Gemini 3.1 Pro
& 50.0
& 67.0 ($+17.0$)
& 67.0 ($+0.0$)
& 34.0
& 36.5 ($+2.5$)
& 46.5 ($+10.0$) \\

Gemini 3.5 Flash
& 48.0
& 62.0 ($+14.0$)
& 64.0 ($+2.0$)
& 34.0
& 39.0 ($+5.0$)
& 47.5 ($+8.5$) \\

\midrule
\rowcolor{gray!15}
Overall
& 44.5
& 58.6 ($+14.1$)
& 60.0 ($+1.4$)
& 34.6
& 34.8 ($+0.2$)
& 41.8 ($+7.0$) \\

\bottomrule
\end{tabular}%
}
%\vspace{-7pt}
\end{table*}

\begin{table*}[!t]
\centering
\caption{\textbf{Offline variant tasks
$\boldsymbol{T_2^*}$ and $\boldsymbol{T_3^*}$.}
The left table compares top-down decomposition from $L_2$ to $L_1$ with
bottom-up abstraction from $L_1$ to $L_2$. The right table quantifies the
accuracy loss from action decoding in cross-horizon consistency.}
\vspace{-7pt}
\label{tab:t2_t3_four_model_ablation}
\setlength{\tabcolsep}{4pt}
\renewcommand{\arraystretch}{1.1}
\captionsetup[subtable]{font=small}

\resizebox{0.87\textwidth}{!}{%
\begin{minipage}{\textwidth}
\centering

\begin{subtable}[t]{0.49\textwidth}
\centering
\resizebox{\linewidth}{!}{%
\begin{tabular}{lcc}
\toprule
\raisebox{0.8ex}{\textbf{Model}} &
\shortstack{\textbf{Top-Down}\\\textbf{Decomposition}} &
\shortstack{\textbf{Bottom-Up}\\\textbf{Abstraction}} \\
\midrule

Qwen2.5-VL-32B-Instruct
& 61.0
& 98.0 ($+37.0$) \\

Qwen3-VL-235B-A22B-Instruct
& 61.5
& 98.0 ($+36.5$) \\

Gemini 3.1 Pro
& 76.5
& 99.0 ($+22.5$) \\

Gemini 3.5 Flash
& 79.0
& 98.5 ($+19.5$) \\

\midrule
\rowcolor{gray!15}
Overall
& 69.5
& 98.4 ($+28.9$) \\

\bottomrule
\end{tabular}%
}
\vspace{2pt}
\caption{\textbf{Goal Decomposition and Abstraction
($\boldsymbol{T_2^*}$)}}
\label{tab:t2_variant}
\end{subtable}
\hfill
\begin{subtable}[t]{0.49\textwidth}
\centering
\resizebox{\linewidth}{!}{%
\begin{tabular}{lcc}
\toprule
\raisebox{0.8ex}{\textbf{Model}} &
\shortstack{$\boldsymbol{L_3\!-\!L_2\!-\!L_1}$\\
\textbf{w/o Actions}} &
\shortstack{$\boldsymbol{L_3\!-\!L_2\!-\!L_1}$\\
\textbf{w/ Actions}} \\
\midrule

Qwen2.5-VL-32B-Instruct
& 69.0
& 68.5 ($-0.5$) \\

Qwen3-VL-235B-A22B-Instruct
& 72.0
& 71.0 ($-1.0$) \\

Gemini 3.1 Pro
& 77.5
& 77.5 ($+0.0$) \\

Gemini 3.5 Flash
& 77.5
& 76.5 ($-1.0$) \\

\midrule
\rowcolor{gray!15}
Overall
& 74.0
& 73.4 ($-0.6$) \\

\bottomrule
\end{tabular}%
}
\vspace{2pt}
\caption{\textbf{Cross-Horizon Consistency
($\boldsymbol{T_3^*}$)}}
\label{tab:t3_variant}
\end{subtable}

\end{minipage}%
}

\vspace{-7pt}
\end{table*}

\noindent \textbf{Multi-Horizon Pyramid.} In \reftab{}~\ref{tab:t1_four_model_ablation}, instruction horizons have distinct effects on current-action perception and future-action planning. For action perception, short-horizon instructions increase average accuracy by $14.1$ points, whereas multi-horizon instructions yield only a further $1.4$-point improvement to $60.0\%$.  Conversely, for future planning, short-horizon instructions produce only a $0.2$-point improvement, while multi-horizon instructions further increase accuracy from $34.8\%$ to $41.8\%$ by $7.0$ points. Short-horizon instructions directly describe immediate operations and therefore align more closely with current actions. Medium- and long-horizon instructions provide sustained goals and game strategies that better guide future planning. These results demonstrate the advantage of our multi-horizon pyramid over the prior simple instructions~\citep{openp2p}. Overall, our multi-horizon framework can support both immediate decision-making and future-task planning.

\noindent \textbf{Top-Down and Bottom-Up.} In \reftab{}~\ref{tab:t2_t3_four_model_ablation}\subref{tab:t2_variant}, all four models perform substantially better on bottom-up abstraction than top-down decomposition. Average accuracy reaches $98.4\%$ for abstraction but only $69.5\%$ for decomposition, with a gap of $28.9$ percentage points. For bottom-up abstraction, concrete low-level action sequences contain sufficient information to identify the corresponding higher-level goals, resulting in near-ceiling accuracy. Although this saturation limits its discriminability for evaluation, the reliable bottom-up mapping is well suited to our annotation pipeline, which constructs higher-level instructions from segmented short-horizon trajectories. In contrast, top-down decomposition starts from abstract goals and strategies that omit operational details and may correspond to multiple valid action sequences. The information insufficiency and underdetermination make top-down decomposition unsuitable for annotation but turn it into a challenging evaluation task for goal decomposition and planning. Together, these results justify our complementary design choices, \textit{i.e.}, the bottom-up abstraction for data annotation and top-down decomposition for model evaluation.

\noindent \textbf{Action-Decoding Error.} In \reftab{}~\ref{tab:t2_t3_four_model_ablation}\subref{tab:t3_variant}, adding action sequences to the cross-horizon consistency task causes only modest performance degradation. Average accuracy decreases from $74.0\%$ without actions to $73.4\%$ with actions, corresponding to a loss of $0.6$ percentage points. The limited gap indicates that decoding low-level action sequences introduces little additional error beyond matching instructions across temporal horizons. Most errors therefore arise from decomposing, planning, and aligning multi-horizon goals rather than from action decoding. The results help to locate the primary bottleneck for current models in gameplay scenarios.

\subsection{Stepwise Online Track}\label{sec:4.4}

\noindent \textbf{Offline--Online Association.} In \reftab{}~\ref{tab:online_results}, online performance closely aligns with offline rankings. Models from offline Tiers~1--4 occupy online ranks 1--3, 4--6, 7--9, and 10--12, respectively. The positive association proves that our offline track captures relevant abilities and provides an effective indicator of online gameplay performance.

\noindent \textbf{Long-Horizon Challenges.} The long-horizon online tasks remain difficult for current models. GPT-6-Astra achieves the highest success rate of $45.0\%$, while ten of the twelve models achieve at most $10.0\%$. Each long-horizon task requires many sequential operations. A single failure in a short-horizon subtask can invalidate the entire task. The results expose existing limitations in long-horizon game planning and strategy formulation.

\noindent \textbf{Stepwise Diagnosis.} As described in \reffig{}~\ref{fig:online-demo} and \refsec{}~\ref{sec:3.4}, each long-horizon causal or thematic task is decomposed into short-horizon subtasks. The stepwise protocol with game resets can localize failures to specific operations. \textit{E.g.}, we can distinguish models that do not know how to craft a cake from those that understand the procedure but perform a certain step incorrectly. This design can support fine-grained diagnosis in long-horizon gameplay.

\begin{table*}[!t]
\centering
\caption{\textbf{Online results of \bench{}.} Due to the high evaluation costs, we test $12$ models in our online track, with three models per offline performance tier. Each entry reports the success rate as a percentage, followed by the number of passed tasks out of the total in parentheses. The offline and online rankings show a clear positive association.}
\vspace{-7pt}
\label{tab:online_results}
\setlength{\tabcolsep}{4pt}
\renewcommand{\arraystretch}{1.1}
\resizebox{\textwidth}{!}{%
\begin{tabular}{llccccc@{\hspace{10pt}}llccccc}
\toprule
\rowcolor{gray!15}
\multicolumn{7}{c}{\textbf{Tier 1}} &
\multicolumn{7}{c}{\textbf{Tier 2}} \\
\midrule

\textbf{Online} &
\multirow{2}{*}{\textbf{Model}} &
\textbf{Offline} &
\textbf{Short-Horizon} &
\textbf{Long-Horizon} &
\multicolumn{1}{c}{\multirow{2}{*}{\textbf{Causal}}} &
\multicolumn{1}{c}{\multirow{2}{*}{\textbf{Thematic}}} &

\textbf{Online} &
\multirow{2}{*}{\textbf{Model}} &
\textbf{Offline} &
\textbf{Short-Horizon} &
\textbf{Long-Horizon} &
\multicolumn{1}{c}{\multirow{2}{*}{\textbf{Causal}}} &
\multicolumn{1}{c}{\multirow{2}{*}{\textbf{Thematic}}} \\

\textbf{Rank} &
&
\textbf{Rank} &
\textbf{Subtasks} &
\textbf{Tasks} &
&
&

\textbf{Rank} &
&
\textbf{Rank} &
\textbf{Subtasks} &
\textbf{Tasks} &
&
\\
\midrule

\medal{gold}{1}
& GPT-6-Astra
& 1
& \textbf{66.1} (41/62)
& \textbf{45.0} (9/20)
& \textbf{40.0} (4/10)
& \textbf{50.0} (5/10)
&
\textbf{4}
& GPT-5.6 Terra
& 14
& 37.1 (23/62)
& 10.0 (2/20)
& 20.0 (2/10)
& 0.0 (0/10) \\

\medal{silver}{2}
& Gemini 3.6 Flash
& 4
& \underline{56.5} (35/62)
& \underline{30.0} (6/20)
& \underline{30.0} (3/10)
& \underline{30.0} (3/10)
&
\textbf{5}
& GPT-5.6 Luna
& 17
& 33.9 (21/62)
& 10.0 (2/20)
& 10.0 (1/10)
& 10.0 (1/10) \\

\medal{bronze}{3}
& Kimi-K3
& 6
& 46.8 (29/62)
& 10.0 (2/20)
& 20.0 (2/10)
& 0.0 (0/10)
&
\textbf{6}
& MiniMax-M3
& 21
& 29.0 (18/62)
& 10.0 (2/20)
& 10.0 (1/10)
& 10.0 (1/10) \\

\midrule
\rowcolor{gray!15}
\multicolumn{7}{c}{\textbf{Tier 3}} &
\multicolumn{7}{c}{\textbf{Tier 4}} \\
\midrule

\textbf{Online} &
\multirow{2}{*}{\textbf{Model}} &
\textbf{Offline} &
\textbf{Short-Horizon} &
\textbf{Long-Horizon} &
\multicolumn{1}{c}{\multirow{2}{*}{\textbf{Causal}}} &
\multicolumn{1}{c}{\multirow{2}{*}{\textbf{Thematic}}} &

\textbf{Online} &
\multirow{2}{*}{\textbf{Model}} &
\textbf{Offline} &
\textbf{Short-Horizon} &
\textbf{Long-Horizon} &
\multicolumn{1}{c}{\multirow{2}{*}{\textbf{Causal}}} &
\multicolumn{1}{c}{\multirow{2}{*}{\textbf{Thematic}}} \\

\textbf{Rank} &
&
\textbf{Rank} &
\textbf{Subtasks} &
\textbf{Tasks} &
&
&

\textbf{Rank} &
&
\textbf{Rank} &
\textbf{Subtasks} &
\textbf{Tasks} &
&
\\
\midrule

\textbf{7}
& GLM-5V-Turbo
& 27
& 27.4 (17/62)
& 5.0 (1/20)
& 10.0 (1/10)
& 0.0 (0/10)
&
\textbf{10}
& Qwen3.6-35B-A3B
& 34
& 11.3 (7/62)
& 5.0 (1/20)
& 10.0 (1/10)
& 0.0 (0/10) \\

\textbf{8}
& Qwen3.5-397B-A17B
& 29
& 19.4 (12/62)
& 5.0 (1/20)
& 10.0 (1/10)
& 0.0 (0/10)
&
\textbf{11}
& InternVL3.5-8B
& 38
& 3.2 (2/62)
& 0.0 (0/20)
& 0.0 (0/10)
& 0.0 (0/10) \\

\textbf{9}
& Step3-VL-10B
& 32
& 14.5 (9/62)
& 5.0 (1/20)
& 10.0 (1/10)
& 0.0 (0/10)
&
\textbf{12}
& UI-TARS-1.5-7B
& 41
& 1.6 (1/62)
& 0.0 (0/20)
& 0.0 (0/10)
& 0.0 (0/10) \\

\bottomrule
\end{tabular}%
}
\vspace{-7pt}
\end{table*}

\section{Conclusion}

{\parfillskip=0pt
We propose \name{}, a unified data and evaluation suite that measures gameplay capabilities at different horizons across diverse model families. First, \anno{} can automatically build a dense instruction pyramid spanning short-horizon operations, medium-horizon goals, and long-horizon strategies. Second, we construct \data{}, a large-scale AAA gameplay dataset with temporally aligned videos, actions, and multi-horizon instructions in $21$ titles. Third, based on our data, \bench{} combines reproducible offline evaluation with stepwise online testing, enabling standardized model comparisons and fine-grained failure diagnosis. Extensive experiments reveal capability differences among current models and identify their primary bottlenecks. Our work can serve as a data foundation and unified yardstick for systematically examining the gameplay performance across model families and temporal horizons.\par
}

%In this paper, we propose \name{}, a unified data and evaluation suite that measures gameplay capabilities at different horizons across diverse model families. Existing gameplay data and benchmarks suffer from narrow game coverage, limited textual instructions, and high-variance online rollouts. To address these challenges, we design three components in our \namesuite{}. First, \anno{} can automatically build a dense instruction pyramid spanning short-horizon operations, medium-horizon goals, and long-horizon strategies. Second, using the annotator, we construct \data{}, a large-scale AAA gameplay dataset with temporally aligned videos, actions, and multi-horizon instructions in $21$ titles. Third, based on our data, \bench{} combines reproducible offline evaluation with stepwise online testing, enabling standardized model comparisons and fine-grained failure diagnosis. Extensive experiments reveal capability differences among current models and identify their primary bottlenecks. Our work can serve as a data foundation and unified yardstick for systematically examining gameplay performance across models and temporal horizons.\par

\clearpage
\bibliographystyle{unsrtnat}
\bibliography{main}

\begin{thebibliography}{50}
\providecommand{\natexlab}[1]{#1}
\providecommand{\url}[1]{\texttt{#1}}
\expandafter\ifx\csname urlstyle\endcsname\relax
  \providecommand{\doi}[1]{doi: #1}\else
  \providecommand{\doi}{doi: \begingroup \urlstyle{rm}\Url}\fi

\bibitem[Magne et~al.(2026)Magne, Awadalla, Wang, Xu, Belofsky, Hu, Kim, Schmidt, Gkioxari, Kautz, Yue, Choi, Zhu, and Fan]{nitrogen}
Lo{\"\i}c Magne, Anas Awadalla, Guanzhi Wang, Yinzhen Xu, Joshua Belofsky, Fengyuan Hu, Joohwan Kim, Ludwig Schmidt, Georgia Gkioxari, Jan Kautz, Yisong Yue, Yejin Choi, Yuke Zhu, and Linxi Fan.
\newblock Nitrogen: An open foundation model for generalist gaming agents.
\newblock In \emph{Proceedings of the IEEE/CVF Conference on Computer Vision and Pattern Recognition (CVPR)}, pages 21511--21521, 2026.

\bibitem[Yue et~al.(2026)Yue, Salia, Hunt, Green, Shi, and Hunt]{openp2p}
Yuguang Yue, Irakli Salia, Samuel Hunt, Chris Green, Wenzhe Shi, and Jonathan~J Hunt.
\newblock Scaling behavior cloning improves causal reasoning: An open model for real-time video game playing.
\newblock \emph{arXiv preprint arXiv:2601.04575}, 2026.

\bibitem[Tan et~al.(2025)Tan, Li, Fang, Yao, Yan, Luo, Ao, Li, Ren, Yi, et~al.]{lumine}
Weihao Tan, Xiangyang Li, Yunhao Fang, Heyuan Yao, Shi Yan, Hao Luo, Tenglong Ao, Huihui Li, Hongbin Ren, Bairen Yi, et~al.
\newblock Lumine: An open recipe for building generalist agents in 3d open worlds.
\newblock \emph{arXiv preprint arXiv:2511.08892}, 2025.

\bibitem[Wang et~al.(2025{\natexlab{a}})Wang, Li, Ye, Fang, Wang, Liu, Liang, Lu, Wu, Feng, et~al.]{gametars}
Zihao Wang, Xujing Li, Yining Ye, Junjie Fang, Haoming Wang, Longxiang Liu, Shihao Liang, Junting Lu, Zhiyong Wu, Jiazhan Feng, et~al.
\newblock Game-tars: Pretrained foundation models for scalable generalist multimodal game agents.
\newblock \emph{arXiv preprint arXiv:2510.23691}, 2025{\natexlab{a}}.

\bibitem[Cai et~al.(2024{\natexlab{a}})Cai, Mu, He, Zhang, Zheng, Liu, and Liang]{minestudio}
Shaofei Cai, Zhancun Mu, Kaichen He, Bowei Zhang, Xinyue Zheng, Anji Liu, and Yitao Liang.
\newblock Minestudio: A streamlined package for minecraft ai agent development.
\newblock \emph{arXiv preprint arXiv:2412.18293}, 2024{\natexlab{a}}.

\bibitem[Comanici et~al.(2025)Comanici, Bieber, Schaekermann, Pasupat, Sachdeva, Dhillon, Blistein, Ram, Zhang, Rosen, et~al.]{gemini25}
Gheorghe Comanici, Eric Bieber, Mike Schaekermann, Ice Pasupat, Noveen Sachdeva, Inderjit Dhillon, Marcel Blistein, Ori Ram, Dan Zhang, Evan Rosen, et~al.
\newblock Gemini 2.5: Pushing the frontier with advanced reasoning, multimodality, long context, and next generation agentic capabilities.
\newblock \emph{arXiv preprint arXiv:2507.06261}, 2025.

\bibitem[{Team Gemini}(2023)]{gemini1}
{Team Gemini}.
\newblock Gemini: a family of highly capable multimodal models.
\newblock \emph{arXiv preprint arXiv:2312.11805}, 2023.

\bibitem[Achiam et~al.(2023)Achiam, Adler, Agarwal, Ahmad, Akkaya, Aleman, Almeida, Altenschmidt, Altman, Anadkat, et~al.]{gpt4}
Josh Achiam, Steven Adler, Sandhini Agarwal, Lama Ahmad, Ilge Akkaya, Florencia~Leoni Aleman, Diogo Almeida, Janko Altenschmidt, Sam Altman, Shyamal Anadkat, et~al.
\newblock Gpt-4 technical report.
\newblock \emph{arXiv preprint arXiv:2303.08774}, 2023.

\bibitem[Baker et~al.(2022)Baker, Akkaya, Zhokov, Huizinga, Tang, Ecoffet, Houghton, Sampedro, and Clune]{vpt}
Bowen Baker, Ilge Akkaya, Peter Zhokov, Joost Huizinga, Jie Tang, Adrien Ecoffet, Brandon Houghton, Raul Sampedro, and Jeff Clune.
\newblock Video pretraining (vpt): Learning to act by watching unlabeled online videos.
\newblock In \emph{Advances in neural information processing systems}, volume~35, pages 24639--24654, 2022.

\bibitem[Bolton et~al.(2025)Bolton, Lerchner, Cordell, Moufarek, Bolt, Lampinen, Mitenkova, Hallingstad, Vujatovic, Li, et~al.]{sima2}
Adrian Bolton, Alexander Lerchner, Alexandra Cordell, Alexandre Moufarek, Andrew Bolt, Andrew Lampinen, Anna Mitenkova, Arne~Olav Hallingstad, Bojan Vujatovic, Bonnie Li, et~al.
\newblock Sima 2: A generalist embodied agent for virtual worlds.
\newblock \emph{arXiv preprint arXiv:2512.04797}, 2025.

\bibitem[Qin et~al.(2025)Qin, Ye, Fang, Wang, Liang, Tian, Zhang, Li, Li, Huang, et~al.]{uitars1}
Yujia Qin, Yining Ye, Junjie Fang, Haoming Wang, Shihao Liang, Shizuo Tian, Junda Zhang, Jiahao Li, Yunxin Li, Shijue Huang, et~al.
\newblock Ui-tars: Pioneering automated gui interaction with native agents.
\newblock \emph{arXiv preprint arXiv:2501.12326}, 2025.

\bibitem[Wang et~al.(2025{\natexlab{b}})Wang, Zou, Song, Feng, Fang, Lu, Liu, Luo, Liang, Huang, et~al.]{uitars2}
Haoming Wang, Haoyang Zou, Huatong Song, Jiazhan Feng, Junjie Fang, Junting Lu, Longxiang Liu, Qinyu Luo, Shihao Liang, Shijue Huang, et~al.
\newblock Ui-tars-2 technical report: Advancing gui agent with multi-turn reinforcement learning.
\newblock \emph{arXiv preprint arXiv:2509.02544}, 2025{\natexlab{b}}.

\bibitem[Wang et~al.(2023)Wang, Xie, Jiang, Mandlekar, Xiao, Zhu, Fan, and Anandkumar]{voyager}
Guanzhi Wang, Yuqi Xie, Yunfan Jiang, Ajay Mandlekar, Chaowei Xiao, Yuke Zhu, Linxi Fan, and Anima Anandkumar.
\newblock Voyager: An open-ended embodied agent with large language models, 2023.
\newblock \emph{arXiv preprint arXiv:2305.16291}, 2023.

\bibitem[Tan et~al.(2024)Tan, Zhang, Xu, Xia, Ding, Li, Zhou, Yue, Jiang, Li, et~al.]{cradle}
Weihao Tan, Wentao Zhang, Xinrun Xu, Haochong Xia, Ziluo Ding, Boyu Li, Bohan Zhou, Junpeng Yue, Jiechuan Jiang, Yewen Li, et~al.
\newblock Cradle: Empowering foundation agents towards general computer control.
\newblock \emph{arXiv preprint arXiv:2403.03186}, 2024.

\bibitem[Li et~al.(2025)Li, Wang, He, Ma, and Liang]{jarvisvla}
Muyao Li, Zihao Wang, Kaichen He, Xiaojian Ma, and Yitao Liang.
\newblock Jarvis-vla: Post-training large-scale vision language models to play visual games with keyboards and mouse.
\newblock In \emph{Findings of the Association for Computational Linguistics: ACL 2025}, pages 17878--17899, 2025.

\bibitem[Cai et~al.(2024{\natexlab{b}})Cai, Zhang, Wang, Ma, Liu, and Liang]{groot}
Shaofei Cai, Bowei Zhang, Zihao Wang, Xiaojian Ma, Anji Liu, and Yitao Liang.
\newblock Groot: Learning to follow instructions by watching gameplay videos.
\newblock In \emph{International Conference on Learning Representations}, 2024{\natexlab{b}}.

\bibitem[Cai et~al.(2025)Cai, Wang, Lian, Mu, Ma, Liu, and Liang]{rocket1}
Shaofei Cai, Zihao Wang, Kewei Lian, Zhancun Mu, Xiaojian Ma, Anji Liu, and Yitao Liang.
\newblock Rocket-1: Mastering open-world interaction with visual-temporal context prompting.
\newblock In \emph{Proceedings of the IEEE/CVF Conference on Computer Vision and Pattern Recognition (CVPR)}, pages 12122--12131, 2025.

\bibitem[Brohan et~al.(2023)Brohan, Brown, Carbajal, Chebotar, Chen, Choromanski, Ding, Driess, Dubey, Finn, et~al.]{RT2}
Anthony Brohan, Noah Brown, Justice Carbajal, Yevgen Chebotar, Xi~Chen, Krzysztof Choromanski, Tianli Ding, Danny Driess, Avinava Dubey, Chelsea Finn, et~al.
\newblock Rt-2: Vision-language-action models transfer web knowledge to robotic control.
\newblock \emph{arXiv preprint arXiv:2307.15818}, 2023.

\bibitem[Kim et~al.(2024)Kim, Pertsch, Karamcheti, Xiao, Balakrishna, Nair, Rafailov, Foster, Lam, Sanketi, et~al.]{openvla}
Moo~Jin Kim, Karl Pertsch, Siddharth Karamcheti, Ted Xiao, Ashwin Balakrishna, Suraj Nair, Rafael Rafailov, Ethan Foster, Grace Lam, Pannag Sanketi, et~al.
\newblock Openvla: An open-source vision-language-action model.
\newblock \emph{arXiv preprint arXiv:2406.09246}, 2024.

\bibitem[{Anthropic}(2025)]{claude37}
{Anthropic}.
\newblock {Claude 3.7 Sonnet and Claude Code}.
\newblock \url{https://www.anthropic.com/news/claude-3-7-sonnet}, February 2025.
\newblock Accessed: 2026-08-20.

\bibitem[Ouyang et~al.(2026)Ouyang, Hu, Lin, Ng, and Shou]{gameworld}
Mingyu Ouyang, Siyuan Hu, Kevin~Qinghong Lin, Hwee~Tou Ng, and Mike~Zheng Shou.
\newblock Gameworld: Towards standardized and verifiable evaluation of multimodal game agents.
\newblock 2026.

\bibitem[Lifshitz et~al.(2023)Lifshitz, Paster, Chan, Ba, and McIlraith]{steve1}
Shalev Lifshitz, Keiran Paster, Harris Chan, Jimmy Ba, and Sheila McIlraith.
\newblock Steve-1: A generative model for text-to-behavior in minecraft.
\newblock 36:\penalty0 69900--69929, 2023.

\bibitem[Fan et~al.(2022)Fan, Wang, Jiang, Mandlekar, Yang, Zhu, Tang, Huang, Zhu, and Anandkumar]{minedojo}
Linxi Fan, Guanzhi Wang, Yunfan Jiang, Ajay Mandlekar, Yuncong Yang, Haoyi Zhu, Andrew Tang, De-An Huang, Yuke Zhu, and Anima Anandkumar.
\newblock Minedojo: Building open-ended embodied agents with internet-scale knowledge.
\newblock 35:\penalty0 18343--18362, 2022.

\bibitem[Li et~al.(2026)Li, Meng, Shi, Peng, Wu, Zheng, Li, and Zhang]{wildworld}
Zhen Li, Zian Meng, Shuwei Shi, Wenshuo Peng, Yuwei Wu, Bo~Zheng, Chuanhao Li, and Kaipeng Zhang.
\newblock Wildworld: A large-scale dataset for dynamic world modeling with actions and explicit state toward generative arpg.
\newblock 2026.

\bibitem[Zhang et~al.(2026)Zhang, Liu, Zhao, Hou, Zhang, Xie, Liu, and Li]{gameverse}
Kuan Zhang, Dongchen Liu, Qiyue Zhao, Jinkun Hou, Xinran Zhang, Qinlei Xie, Miao Liu, and Yiming Li.
\newblock Gameverse: Can vision-language models learn from video-based reflection?
\newblock \emph{arXiv preprint arXiv:2603.06656}, 2026.

\bibitem[Choi et~al.(2026)Choi, Jung, Seong, Kim, Kim, Cho, Kim, Park, Yu, and Lee]{d2e}
Suhwan Choi, Jaeyoon Jung, Haebin Seong, Minchan Kim, Minyeong Kim, Yongjun Cho, Yoonshik Kim, Yu~Park, Youngjae Yu, and Yunsung Lee.
\newblock D2e: Scaling vision-action pretraining on desktop data for transfer to embodied ai.
\newblock In \emph{International Conference on Learning Representations}, 2026.

\bibitem[{Markov AI}(2026)]{gaming500h}
{Markov AI}.
\newblock gaming-500-hours.
\newblock Hugging Face Datasets, \url{https://huggingface.co/datasets/markov-ai/gaming-500-hours}, 2026.
\newblock Accessed: 2026-08-20.

\bibitem[Bai et~al.(2026)Bai, Bai, Bao, Cai, Cai, Cao, Cao, Chai, Charles, et~al.]{kimik3}
Tongtong Bai, Yifan Bai, Yiping Bao, Jianfeng Cai, Xinyuan Cai, Peizhou Cao, Yuxuan Cao, Ziwei Chai, Y~Charles, et~al.
\newblock Kimi k3: Open frontier intelligence.
\newblock \emph{arXiv preprint arXiv:2607.24653}, 2026.

\bibitem[Yang et~al.(2025)Yang, Li, Yang, Zhang, Hui, Zheng, Yu, Gao, Huang, Lv, et~al.]{qwen3}
An~Yang, Anfeng Li, Baosong Yang, Beichen Zhang, Binyuan Hui, Bo~Zheng, Bowen Yu, Chang Gao, Chengen Huang, Chenxu Lv, et~al.
\newblock Qwen3 technical report.
\newblock \emph{arXiv preprint arXiv:2505.09388}, 2025.

\bibitem[Diao et~al.(2026)Diao, Wu, Deng, Wang, Bai, Wu, Fan, Ye, Tong, Fan, et~al.]{sensenova}
Haiwen Diao, Penghao Wu, Hanming Deng, Jiahao Wang, Shihao Bai, Silei Wu, Weichen Fan, Wenjie Ye, Wenwen Tong, Xiangyu Fan, et~al.
\newblock Sensenova-u1: Unifying multimodal understanding and generation with neo-unify architecture.
\newblock \emph{arXiv preprint arXiv:2605.12500}, 2026.

\bibitem[Wang et~al.(2025{\natexlab{c}})Wang, Zhao, Zhang, Cao, Zhan, Duan, Lu, Fu, Chen, Zhao, et~al.]{ovis}
Guo-Hua Wang, Shanshan Zhao, Xinjie Zhang, Liangfu Cao, Pengxin Zhan, Lunhao Duan, Shiyin Lu, Minghao Fu, Xiaohao Chen, Jianshan Zhao, et~al.
\newblock Ovis-u1 technical report.
\newblock \emph{arXiv preprint arXiv:2506.23044}, 2025{\natexlab{c}}.

\bibitem[Tian et~al.(2026)Tian, Yang, Chen, Cui, Wang, Duan, Yin, Chen, Yang, Liu, et~al.]{internvl-u}
Changyao Tian, Danni Yang, Guanzhou Chen, Erfei Cui, Zhaokai Wang, Yuchen Duan, Penghao Yin, Sitao Chen, Ganlin Yang, Mingxin Liu, et~al.
\newblock Internvl-u: Democratizing unified multimodal models for understanding, reasoning, generation and editing.
\newblock \emph{arXiv preprint arXiv:2603.09877}, 2026.

\bibitem[Deng et~al.(2025)Deng, Zhu, Li, Gou, Li, Wang, Zhong, Yu, Nie, Song, et~al.]{bagel}
Chaorui Deng, Deyao Zhu, Kunchang Li, Chenhui Gou, Feng Li, Zeyu Wang, Shu Zhong, Weihao Yu, Xiaonan Nie, Ziang Song, et~al.
\newblock Emerging properties in unified multimodal pretraining.
\newblock \emph{arXiv preprint arXiv:2505.14683}, 2025.

\bibitem[{GELab-Team, StepFun}(2025)]{stepgui}
{GELab-Team, StepFun}.
\newblock Step-{GUI} technical report.
\newblock \emph{arXiv preprint arXiv:2512.15431}, 2025.

\bibitem[Anthropic(2026)]{claude}
Anthropic.
\newblock Claude model system cards.
\newblock \url{https://www.anthropic.com/system-cards}, 2026.
\newblock Accessed: 2026-08-20.

\bibitem[Guss et~al.(2019)Guss, Houghton, Topin, Wang, Codel, Veloso, and Salakhutdinov]{minerl}
William~H. Guss, Brandon Houghton, Nicholay Topin, Phillip Wang, Cayden Codel, Manuela Veloso, and Ruslan Salakhutdinov.
\newblock Minerl: a large-scale dataset of minecraft demonstrations.
\newblock In \emph{Proceedings of the 28th International Joint Conference on Artificial Intelligence}, page 2442–2448, 2019.

\bibitem[Zheng et~al.(2025)Zheng, Lin, He, Wang, Fu, Fu, Zheng, and Liang]{mcu}
Xinyue Zheng, Haowei Lin, Kaichen He, Zihao Wang, Qiang Fu, Haobo Fu, Zilong Zheng, and Yitao Liang.
\newblock {MCU}: An evaluation framework for open-ended game agents.
\newblock In \emph{International conference on machine learning}, pages 78221--78259. PMLR, 2025.

\bibitem[Zhang et~al.(2025)Zhang, Griffiths, Narasimhan, and Press]{videogamebench}
Alex~L Zhang, Thomas~L Griffiths, Karthik~R Narasimhan, and Ofir Press.
\newblock Videogamebench: Can vision-language models complete popular video games?
\newblock \emph{arXiv preprint arXiv:2505.18134}, 2025.

\bibitem[Tan and Le(2019)]{efficientnet}
Mingxing Tan and Quoc Le.
\newblock {E}fficient{N}et: Rethinking model scaling for convolutional neural networks.
\newblock In \emph{International conference on machine learning}, pages 6105--6114. PMLR, 2019.

\bibitem[Xie et~al.(2021)Xie, Wang, Yu, Anandkumar, Alvarez, and Luo]{segformer}
Enze Xie, Wenhai Wang, Zhiding Yu, Anima Anandkumar, Jose~M Alvarez, and Ping Luo.
\newblock Segformer: Simple and efficient design for semantic segmentation with transformers.
\newblock 34:\penalty0 12077--12090, 2021.

\bibitem[Castellano(2024)]{pyscene}
Brandon Castellano.
\newblock {PySceneDetect}.
\newblock \url{https://www.scenedetect.com}, 2024.
\newblock Software; accessed August 20, 2026.

\bibitem[{Google DeepMind}(2026{\natexlab{a}})]{gemini35flash}
{Google DeepMind}.
\newblock {Gemini 3.5 Flash}: Model card.
\newblock \url{https://deepmind.google/models/model-cards/gemini-3-5-flash/}, 2026{\natexlab{a}}.
\newblock Accessed: 2026-09-05.

\bibitem[Bellman(1966)]{bellman1966dynamic}
Richard Bellman.
\newblock Dynamic {P}rogramming.
\newblock \emph{Science}, 153\penalty0 (3731):\penalty0 34--37, 1966.

\bibitem[He et~al.(2025)He, Weilbach, Wojciechowska, Zhang, and Wood]{plaicraft}
Yingchen He, Christian~D Weilbach, Martyna~E Wojciechowska, Yuxuan Zhang, and Frank Wood.
\newblock Plaicraft: Large-scale time-aligned vision-speech-action dataset for embodied ai.
\newblock \emph{arXiv preprint arXiv:2505.12707}, 2025.

\bibitem[Guo et~al.(2026)Guo, Liang, Liu, Liu, Huang, Hancke, and Lau]{egocs400}
Rongjin Guo, Dong Liang, Yuhao Liu, Fang Liu, Tianyu Huang, Gerhard~P Hancke, and Rynson~WH Lau.
\newblock Egocs-400k: An egocentric gameplay dataset for world models.
\newblock \emph{arXiv preprint arXiv:2606.18180}, 2026.

\bibitem[{Google DeepMind}(2026{\natexlab{b}})]{gemini36flash}
{Google DeepMind}.
\newblock {Gemini 3.6 Flash}: Model card.
\newblock \url{https://deepmind.google/models/model-cards/gemini-3-6-flash/}, 2026{\natexlab{b}}.
\newblock Accessed: 2026-09-05.

\bibitem[Ju et~al.(2026)Ju, Sun, Wu, Zhang, Huo, Su, Gu, Cai, Liu, and Zhang]{mineexp}
Tianjie Ju, Yueqing Sun, Zheng Wu, Wei Zhang, Yaqi Huo, Xi~Su, Qi~Gu, Xunliang Cai, Gongshen Liu, and Zhuosheng Zhang.
\newblock Mineexplorer: Evaluating open-world exploration of mllm agents in minecraft, 2026.

\bibitem[Paglieri et~al.(2025)Paglieri, Cupia\l, Coward, Piterbarg, Wo\l~czyk, Khan, Pignatelli, Kuci\'{n}ski, Pinto, Fergus, Foerster, Parker-Holder, and Rocktaeschel]{balrog}
Davide Paglieri, Bart\l~omiej Cupia\l, Samuel Coward, Ulyana Piterbarg, Maciej Wo\l~czyk, Akbir Khan, Eduardo Pignatelli, \L~ukasz Kuci\'{n}ski, Lerrel Pinto, Rob Fergus, Jakob Foerster, Jack Parker-Holder, and Tim Rocktaeschel.
\newblock {BALROG}: Benchmarking agentic {LLM} and {VLM} reasoning on games.
\newblock In \emph{International Conference on Learning Representations}, 2025.

\bibitem[Park et~al.(2026)Park, Kim, Choi, Kim, Lee, Lee, Park, Lee, Hwang, AHN, Mahabaleshwarkar, Kartal, Biswas, Suhara, Lee, and Cho]{orak}
Dongmin Park, Minkyu Kim, Beongjun Choi, Junhyuck Kim, Keon Lee, Jonghyun Lee, Inkyu Park, ByeongUk Lee, Jaeyoung Hwang, JAEWOO AHN, Ameya Mahabaleshwarkar, Bilal Kartal, Pritam Biswas, Yoshi Suhara, Kangwook Lee, and Jaewoong Cho.
\newblock Orak: A foundational benchmark for training and evaluating {LLM} agents on diverse video games.
\newblock In \emph{International Conference on Learning Representations}, 2026.

\bibitem[Needleman and Wunsch(1970)]{needleman1970general}
Saul~B Needleman and Christian~D Wunsch.
\newblock A general method applicable to the search for similarities in the amino acid sequence of two proteins.
\newblock \emph{Journal of molecular biology}, 48\penalty0 (3):\penalty0 443--453, 1970.

\end{thebibliography}

\clearpage
\beginappendix
This appendix contains the following contents:
\begin{itemize}
    \addtolength{\leftskip}{1.2em}
    \item[-] More details on our \anno{}, \textit{e.g.}, annotation prompts, segmentation, and keybinds.
    %\item[-] More demos and examples of our \data{}, e.g., the aligned videos, actions, and instructions.
    \item[-] More experimental results on our \bench{}, e.g., qualitative examples of the online track.
\end{itemize}

\section{More Details on \anno{}}

\subsection{Annotation Prompts}
\begin{figure}[!t]
    \centering
    \vspace{-1pt}  
    \includegraphics[width=\textwidth,trim=0 0 0 0,clip]{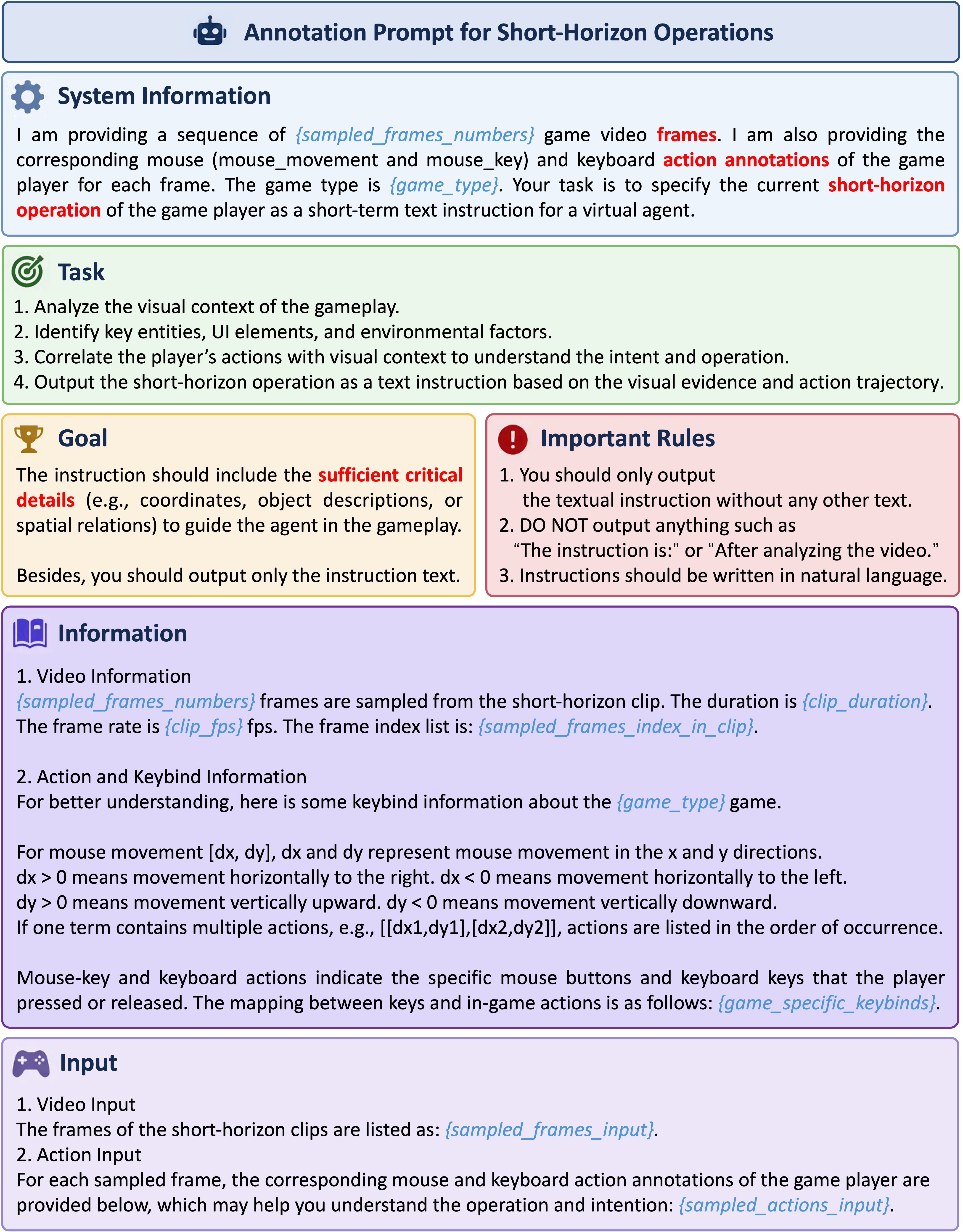}
    \vspace{-14pt}
    \caption{\textbf{The prompt for short-horizon operations.} Key differences across horizons are highlighted in red, such as the inputs, outputs, and required levels of detail in the instructions. Blue text denotes the code variables in the prompt.  The short-horizon prompt takes sampled frames and temporally aligned keyboard-mouse actions as inputs. It requires the VLM to describe the current operation with sufficient details, such as the object descriptions and spatial relations.}

  \label{fig:sp} 
  \vspace{-7pt}
\end{figure}

\begin{figure}[!t]
    \centering
    \vspace{-1pt}  
    \includegraphics[width=\textwidth,trim=0 0 0 0,clip]{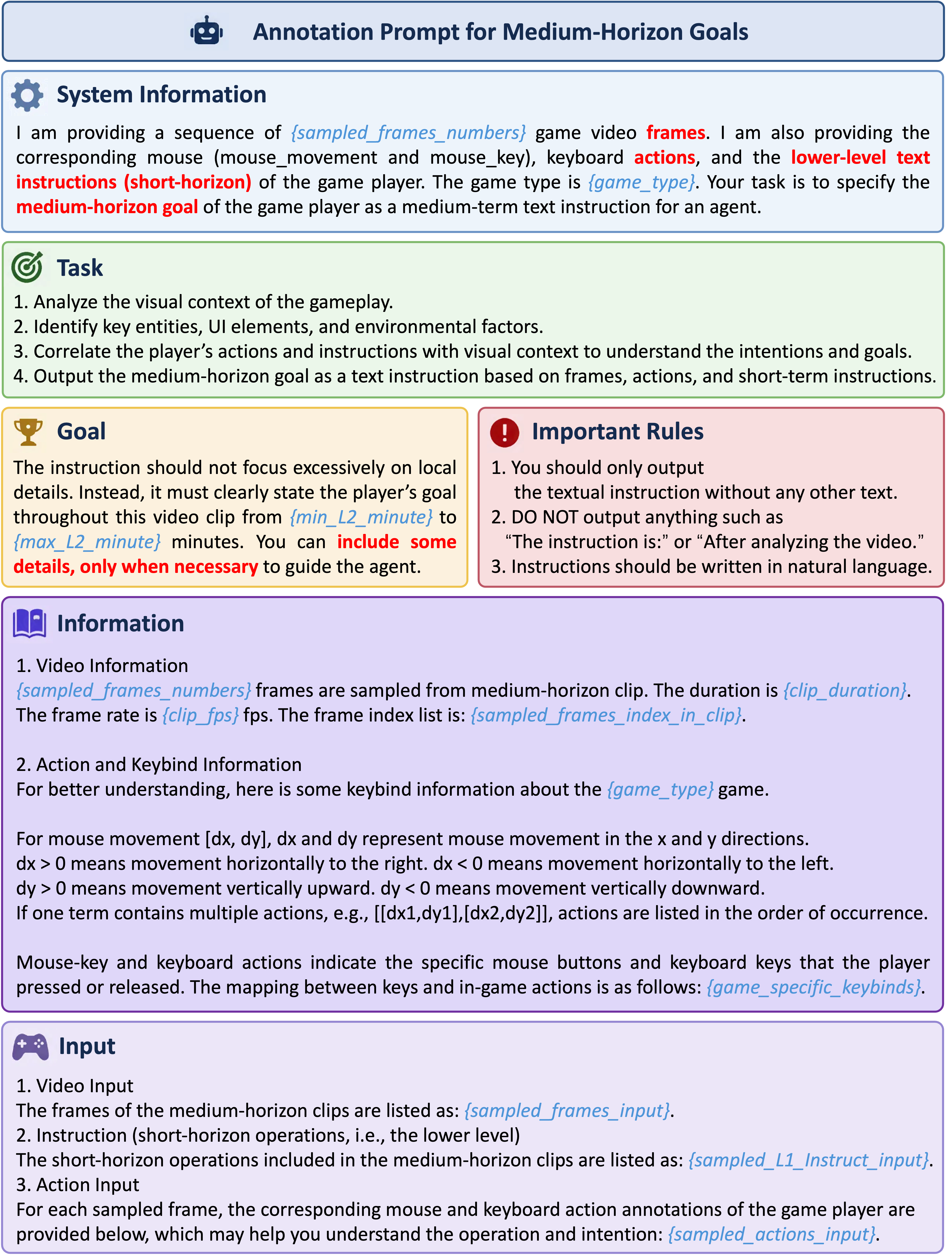}
    \vspace{-14pt}
    \caption{\textbf{The prompt for medium-horizon goals.} Key differences across the three horizons are highlighted in red, such as the inputs, outputs, and required levels of detail in the instructions. Blue text denotes the code variables in the prompt. The medium-horizon prompt incorporates video frames, aligned actions, and the constituent short-horizon instructions, requiring the VLM to summarize the sustained goal of the player while avoiding excessive local details.}

  \label{fig:mp} 
  \vspace{-7pt}
\end{figure}

\begin{figure}[!t]
    \centering
    \vspace{-1pt}  
    \includegraphics[width=\textwidth,trim=0 0 0 0,clip]{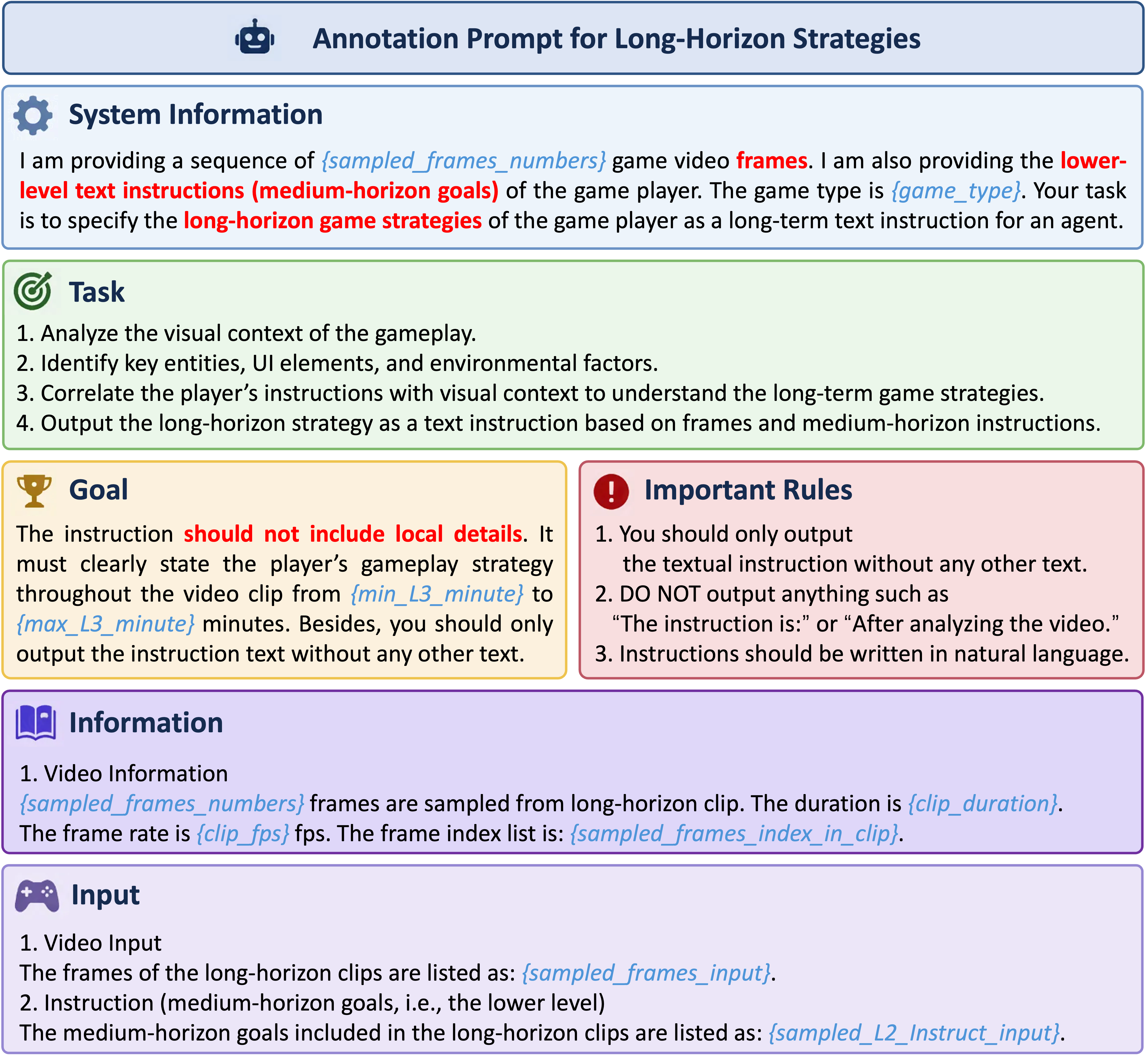}
    \vspace{-14pt}
    \caption{\textbf{The prompt for long-horizon strategies.} Key differences across three horizons are highlighted in red, such as the inputs, outputs, and required levels of detail in the instructions. Blue text denotes the code variables in the prompt. The long-horizon prompt takes sampled frames and the constituent medium-horizon goals as inputs while omitting actions and keybinds. It requires the VLM to capture the overall gameplay strategy without local operational details.}

  \label{fig:lp} 
  \vspace{-7pt}
\end{figure}

As described in \refsec{}~\ref{sec:3.2} of the main paper, we utilize a VLM to annotate the multi-horizon instructions. Here, we present the annotation prompts for short-horizon operations, medium-horizon goals, and long-horizon strategies in \reffig{}~\ref{fig:sp}, \reffig{}~\ref{fig:mp}, and \reffig{}~\ref{fig:lp}, respectively. The short-horizon prompt takes sampled frames and temporally aligned keyboard-mouse actions as inputs. It requires the VLM to describe the current operation with sufficient critical details, such as object descriptions and spatial relations. The medium-horizon prompt also incorporates the constituent short-horizon instructions and asks the VLM to summarize the player's goal while avoiding excessive local details. Game-specific keybinds are additionally provided in the short- and medium-horizon prompts to help the VLM interpret keyboard and mouse actions in terms of their in-game semantics. The long-horizon prompt takes sampled frames and the constituent medium-horizon goals as inputs while omitting actions and keybinds. It requires the VLM to capture the player's overall gameplay strategy without local operational details. Across all three horizons, the prompts provide common video metadata, including the game type, clip duration, frame rate, and sampled frame indices. They also require the VLM to analyze the visual context and output only a natural-language instruction without any additional explanation.

\subsection{Action-Aware Video Segmentation}

{\parfillskip=0pt
We provide more details on the action-aware video segmentation. The segmentation follows four stages, including keyboard-mouse semantic calibration, primitive action classification, rule-based coarse segmentation, and VLM-based refinement. First, using game-specific keybinds, we map raw keyboard and mouse events to gameplay actions, such as movement, attack, interaction, skill activation, and menu operations. When one input can correspond to multiple actions, we retain all candidates instead of making an early decision. Second, we classify primitive actions as input-deterministic, input-ambiguous, or vision-dependent according to how reliably their semantics can be inferred from the inputs. Third, we replay the frame-aligned input logs to recover held keys, discrete triggers, mouse-button states, and camera movements. When multiple actions occur in the same frame, action-priority rules select the primary label, while the remaining actions are retained as concurrent evidence. Consecutive frames with identical labels are merged, short noisy segments are smoothed, and overly long segments are divided. Finally, ambiguous or vision-dependent segments are refined by a VLM using videos, candidate labels, action semantics, and trigger timestamps. We further remove temporal gaps and overlaps, producing segments with continuous action events. Our \anno{} is also compatible with segmentation tools such as PySceneDetect~\citep{pyscene}, which detects boundaries based on inter-frame visual similarity and can therefore fragment continuous actions under large camera movements.\par
}

\subsection{Game-Specific Keybinds}

{\parfillskip=0pt
As mentioned in \refsec{}~\ref{sec:3.2} and \refsec{}~\ref{sec:3.4} of the main manuscript, we adopt the game-specific keybinds for mapping the raw keyboard and mouse inputs to in-game action semantics. We showcase the keybinds for Apex Legends and Cyberpunk 2077 as two representative examples in \reftab{}~\ref{tab:apex_keybinds} and \reftab{}~\ref{tab:cyberpunk_keybinds}, respectively.\par
}

\begin{table*}[!t]
\centering
\caption{\textbf{The keybind for Apex Legends.}
Keyboard and mouse inputs are mapped to their corresponding in-game
action semantics via game-specific keybinds. Multiple candidate semantics for an ambiguous input are separated by slashes.}
\vspace{-7pt}
\label{tab:apex_keybinds}
\setlength{\tabcolsep}{8pt}
\renewcommand{\arraystretch}{1.1}
\resizebox{0.77\textwidth}{!}{%
\begin{tabular}{ll}
\toprule
\textbf{Input Event} & \textbf{In-Game Action Semantics} \\
\midrule

\texttt{KEY\_W} & Move Forward \\
\texttt{KEY\_S} & Move Back \\
\texttt{KEY\_A} & Move Left \\
\texttt{KEY\_D} & Move Right \\
\texttt{SPACE} & Jump \\
\texttt{LSHIFT} & Sprint \\
\texttt{KEY\_C} & Crouch -- Toggle \\
\texttt{LCONTROL} & Crouch -- Hold \\
\texttt{RI\_MOUSE\_LEFT\_BUTTON\_DOWN} & Attack \\
\texttt{RI\_MOUSE\_RIGHT\_BUTTON\_DOWN} & Aim Down Sight \\
\texttt{KEY\_R} & Reload \\
\texttt{KEY\_V} & Melee \\
\texttt{KEY\_1} & Equip Weapon 1 \\
\texttt{KEY\_2} & Equip Weapon 2 \\
\texttt{KEY\_3} & Holster Weapons \\
\texttt{RI\_MOUSE\_WHEEL} & Cycle Weapon \\
\texttt{KEY\_G} & Equip Grenade \\
\texttt{KEY\_N} & Inspect Weapon \\
\texttt{KEY\_Q} & Tactical Ability \\
\texttt{KEY\_Z} & Ultimate Ability \\
\texttt{KEY\_E} & Interact / Pickup / Open Door / Revive / Use Zipline \\
\texttt{KEY\_4} & Use Selected Health Item \\
\texttt{KEY\_H} & Character Utility Action \\
\texttt{RI\_MOUSE\_MIDDLE\_BUTTON\_DOWN} &
Smart Tag / Middle Mouse Button \\
\texttt{TAB} & Open/Close Inventory \\
\texttt{KEY\_M} & Open/Close Map \\

\bottomrule
\end{tabular}%
}
\vspace{-7pt}
\end{table*}

\begin{table*}[!t]
\centering
\caption{\textbf{The keybind for Cyberpunk 2077.}
Keyboard and mouse inputs are mapped to their corresponding in-game action
semantics via game-specific keybinds. Multiple candidate semantics for an
ambiguous input are separated by slashes.}
\vspace{-7pt}
\label{tab:cyberpunk_keybinds}
\setlength{\tabcolsep}{8pt}
\renewcommand{\arraystretch}{1.1}
\resizebox{0.77\textwidth}{!}{%
\begin{tabular}{ll}
\toprule
\textbf{Input Event} & \textbf{In-Game Action Semantics} \\
\midrule

\texttt{KEY\_W} & Move Forward / Vehicle Accelerate \\
\texttt{KEY\_S} & Move Back / Vehicle Brake or Reverse \\
\texttt{KEY\_A} & Move Left / Vehicle Turn Left \\
\texttt{KEY\_D} & Move Right / Vehicle Turn Right \\
\texttt{SPACE} & Jump / Vehicle Handbrake / Pause Braindance \\
\texttt{LSHIFT} & Sprint -- Toggle / Change Braindance Layer \\
\texttt{RSHIFT} & Sprint -- Hold \\
\texttt{KEY\_C} & Crouch -- Toggle / Skip Dialogue \\
\texttt{LCONTROL} & Crouch -- Hold / Vehicle Horn / Panzer Smokescreen \\
\texttt{RI\_MOUSE\_LEFT\_BUTTON\_DOWN} &
Shoot / Fast Attack / Panzer Primary Cannon \\
\texttt{RI\_MOUSE\_RIGHT\_BUTTON\_DOWN} &
Aim / Block / Panzer Missile Launcher \\
\texttt{RI\_MOUSE\_MIDDLE\_BUTTON\_DOWN} &
Use Combat Gadget / Tag / Reverse Camera / Braindance Zoom \\
\texttt{RI\_MOUSE\_WHEEL} &
Cycle Weapon / Next or Previous Weapon / Braindance Zoom \\
\texttt{KEY\_R} & Reload / Vehicle Radio / Restart Braindance \\
\texttt{KEY\_Q} & Quick Attack / Vehicle Cycle Camera / Rewind Braindance \\
\texttt{KEY\_E} & Cyberware Systems / Fast-Forward Braindance \\
\texttt{KEY\_F} & Interact / Pickup / Enter or Exit Vehicle \\
\texttt{KEY\_X} & Use Consumable / Exit Braindance \\
\texttt{KEY\_Y} & Skip Dialogue / Dialogue Confirm \\
\texttt{KEY\_Z} & Open Notifications / Quickhack Details \\
\texttt{KEY\_T} & Open Phone \\
\texttt{TAB} & Scan -- Hold / Braindance Analysis Mode \\
\texttt{KEY\_1} & Select First Weapon / Previous Item \\
\texttt{KEY\_2} & Select Second Weapon \\
\texttt{KEY\_3} & Select Third Weapon / Next Item \\
\texttt{LMENU} & Cycle Weapon / Weapon Wheel / Vehicle Cycle Lights \\
\texttt{KEY\_V} & Call Vehicle \\
\texttt{KEY\_B} & Draw or Withdraw Weapon \\
\texttt{KEY\_I} & Open Main Menu / Inventory \\
\texttt{KEY\_M} & Open Map \\
\texttt{KEY\_J} & Open Journal \\
\texttt{KEY\_K} & Open Crafting \\
\texttt{ESCAPE} & Pause Menu / Close Menu \\

\bottomrule
\end{tabular}%
}
\vspace{-7pt}
\end{table*}

% {
% \small
% \setlength{\tabcolsep}{5pt}
% \renewcommand{\arraystretch}{1.1}
% \setlength{\LTleft}{\fill}
% \setlength{\LTright}{\fill}

\section{More Results on \bench{}}

\begin{figure}[!t]
    \centering
    \vspace{-1pt}  
    \includegraphics[width=\textwidth,trim=0 0 0 0,clip]{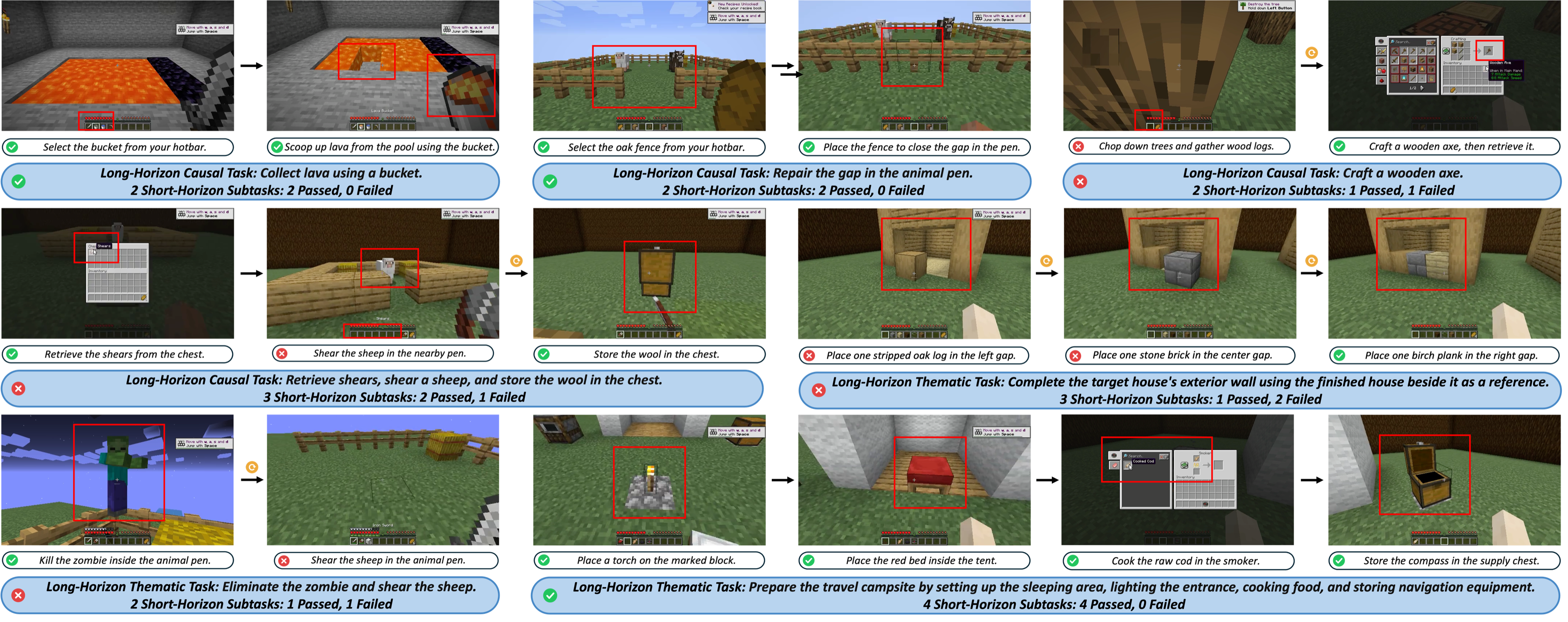}
    \vspace{-14pt}
    \caption{\textbf{The online track of \bench{}.} It contains long-horizon causal and thematic tasks, each comprising multiple verifiable short-horizon subtasks. The causal tasks can only be completed in a prescribed order because of dependencies between successive subtasks. The thematic tasks contain subtasks that share a common theme but can be performed in any order. The figure shows results from Gemini 3.6 Flash~\citep{gemini36flash}, with arrows indicating its actual execution order. After a subtask fails, the environment is reset to the corresponding success state so that evaluation can continue. A long-horizon task is considered passed only when all constituent subtasks succeed.}

  \label{fig:supp_online} 

\end{figure}

We present more results from the online track of our benchmark. In \reffig{}~\ref{fig:supp_online}, we showcase rollouts of Gemini 3.6 Flash~\citep{gemini36flash}, with arrows indicating its actual execution order. In \reftab{}~\ref{tab:online_task_details}, we list all $20$ long-horizon tasks and $62$ constituent short-horizon subtasks, grouped into the causal and thematic categories.

\clearpage

{
\small
\setlength{\tabcolsep}{4pt}
\renewcommand{\arraystretch}{1.1}
\setlength{\LTleft}{\fill}
\setlength{\LTright}{\fill}

\def\taskitem#1#2{%
  \noindent
  \hangindent=2.2em
  \hangafter=1
  \makebox[2.2em][l]{(#1)}#2\par
}

\begin{longtable}{
l
>{\raggedright\arraybackslash}p{0.24\textwidth}
c
p{0.57\textwidth}
}

\caption{\textbf{Long-horizon online tasks and the constituent short-horizon subtasks.}
The online track contains $20$ long-horizon tasks and $62$ short-horizon subtasks.
The causal tasks impose a fixed order, whereas thematic tasks allow any order.}
\label{tab:online_task_details} \\

\toprule
\textbf{No.} &
\textbf{Long-Horizon Task} &
\textbf{Steps} &
\multicolumn{1}{c}{\textbf{Short-Horizon Subtasks}} \\
\midrule
\endfirsthead

\multicolumn{4}{c}{
\textbf{\tablename~\thetable{} continued}
} \\
\toprule
\textbf{No.} &
\textbf{Long-Horizon Task} &
\textbf{Steps} &
\multicolumn{1}{l}{\textbf{Short-Horizon Subtasks}} \\
\midrule
\endhead

\midrule
\multicolumn{4}{r}{\textit{Continued on the next page}} \\
\endfoot

\bottomrule
\endlastfoot

\rowcolor{gray!15}
\multicolumn{4}{c}{\textbf{Causal Tasks}} \\
\midrule

1
& Craft a wooden axe.
& 2
& \taskitem{1}{Chop down trees and gather wood logs outside.}
  \taskitem{2}{Use the crafting table to craft a wooden axe and retrieve it.}
\\[2pt]
\cmidrule(lr){1-4}

2
& Craft an armor stand from workshop materials.
& 5
& \taskitem{1}{Craft four oak planks from an oak log.}
  \taskitem{2}{Craft eight sticks from the oak planks.}
  \taskitem{3}{Smelt three stone blocks into smooth stone.}
  \taskitem{4}{Craft six smooth stone slabs.}
  \taskitem{5}{Craft an armor stand from the sticks and a smooth stone slab.}
\\[2pt]
\cmidrule(lr){1-4}

3
& Craft an item frame from workshop materials.
& 3
& \taskitem{1}{Craft eight oak planks from two oak logs.}
  \taskitem{2}{Craft eight sticks from four oak planks.}
  \taskitem{3}{Craft an item frame from eight sticks and one leather.}
\\[2pt]
\cmidrule(lr){1-4}

4
& Retrieve shears, shear a sheep, and store the wool in the chest.
& 3
& \taskitem{1}{Retrieve the shears from the chest.}
  \taskitem{2}{Shear the sheep in the nearby pen.}
  \taskitem{3}{Store the wool in the chest.}
\\[2pt]
\cmidrule(lr){1-4}

5
& Fill the remaining bucket with milk, bake a cake, and place it on the dining table.
& 3
& \taskitem{1}{Fill the remaining empty bucket with milk from the cow.}
  \taskitem{2}{Craft a cake at the crafting table.}
  \taskitem{3}{Place the cake on the marked dining table.}
\\[2pt]
\cmidrule(lr){1-4}

6
& Repair the gap in the animal pen.
& 2
& \taskitem{1}{Select the oak fence from the hotbar.}
  \taskitem{2}{Place the oak fence to close the gap in the pen.}
\\[2pt]
\cmidrule(lr){1-4}

7
& Mine the missing obsidian block, repair the ruined portal, and ignite it.
& 3
& \taskitem{1}{Mine the loose obsidian block using the diamond pickaxe.}
  \taskitem{2}{Place the obsidian in the missing portal-frame position.}
  \taskitem{3}{Ignite the completed Nether portal with flint and steel.}
\\[2pt]
\cmidrule(lr){1-4}

8
& Collect lava using a bucket.
& 2
& \taskitem{1}{Select the bucket from the hotbar.}
  \taskitem{2}{Scoop up lava from the pool using the empty bucket.}
\\

9
& Mine iron, forge an iron sword, and use it to defeat the zombie.
& 4
& \taskitem{1}{Mine two iron ore blocks using the stone pickaxe.}
  \taskitem{2}{Smelt the iron ore into two iron ingots.}
  \taskitem{3}{Craft an iron sword.}
  \taskitem{4}{Kill the zombie using the iron sword.}
\\[2pt]
\cmidrule(lr){1-4}

10
& Select the golden sword and use it to kill the spider.
& 2
& \taskitem{1}{Select the golden sword from the hotbar.}
  \taskitem{2}{Kill the spider using the golden sword.}
\\[4pt]

\midrule
\rowcolor{gray!15}
\multicolumn{4}{c}{\textbf{Thematic Tasks}} \\
\midrule

1
& Complete the target house's exterior wall using the finished house beside it as a reference.
& 6
& \taskitem{1}{Place a stripped oak log in the ground-row left gap.}
  \taskitem{2}{Place a stone brick block in the ground-row center gap.}
  \taskitem{3}{Place a birch plank in the ground-row right gap.}
  \taskitem{4}{Place a dark oak plank in the second-row left gap.}
  \taskitem{5}{Place a glass block in the second-row center gap.}
  \taskitem{6}{Place a spruce plank in the second-row right gap.}
\\[2pt]
\cmidrule(lr){1-4}

2
& Eliminate the zombie and shear the sheep.
& 2
& \taskitem{1}{Kill the zombie inside the animal pen.}
  \taskitem{2}{Shear the sheep in the animal pen.}
\\[2pt]
\cmidrule(lr){1-4}

3
& Store items from the inventory into the chest.
& 2
& \taskitem{1}{Store the iron ingots in the chest.}
  \taskitem{2}{Store the bread in the chest.}
\\[2pt]
\cmidrule(lr){1-4}

4
& Assemble a mixed-material toolkit at the workshop.
& 3
& \taskitem{1}{Craft an iron pickaxe.}
  \taskitem{2}{Craft a wooden axe.}
  \taskitem{3}{Craft a diamond sword.}
\\[2pt]
\cmidrule(lr){1-4}

5
& Secure and light the tunnel to mine coal and iron.
& 4
& \taskitem{1}{Fill a gap in the passage using cobblestone.}
  \taskitem{2}{Place a torch in the mine.}
  \taskitem{3}{Mine coal ore.}
  \taskitem{4}{Mine iron ore.}
\\[2pt]
\cmidrule(lr){1-4}

6
& Complete routine farm chores by gathering food, planting crops, and managing livestock.
& 5
& \taskitem{1}{Milk the cow using an empty bucket.}
  \taskitem{2}{Pick up the egg lying on the ground.}
  \taskitem{3}{Plant wheat seeds on the farmland.}
  \taskitem{4}{Kill the sheep using the diamond sword.}
  \taskitem{5}{Kill the chicken using the iron sword.}
\\[2pt]
\cmidrule(lr){1-4}

7
& Collect raw blocks for building.
& 2
& \taskitem{1}{Dig dirt blocks using an iron shovel.}
  \taskitem{2}{Mine stone blocks using a diamond pickaxe.}
\\[2pt]
\cmidrule(lr){1-4}

8
& Process building materials and food at the shelter.
& 2
& \taskitem{1}{Smelt cobblestone in the furnace.}
  \taskitem{2}{Cook raw mutton in the smoker.}
\\[2pt]
\cmidrule(lr){1-4}

9
& Assemble a wooden toolkit at the workshop.
& 3
& \taskitem{1}{Craft a wooden pickaxe.}
  \taskitem{2}{Craft a wooden axe.}
  \taskitem{3}{Craft a wooden shovel.}
\\[2pt]
\cmidrule(lr){1-4}

10
& Prepare the travel campsite by setting up the sleeping area, lighting the entrance,
cooking food, and storing navigation equipment.
& 4
& \taskitem{1}{Place the red bed on the wooden sleeping pad inside the tent.}
  \taskitem{2}{Place a torch on the marked block at the campsite entrance.}
  \taskitem{3}{Cook the raw cod in the smoker.}
  \taskitem{4}{Store the compass in the supply chest.}
\\

\end{longtable}
}

\end{document}